\documentclass[lettersize,journal]{IEEEtran}
\usepackage{amsmath,amsfonts}
\usepackage{algorithm}
\usepackage{algorithmic}
\usepackage{multirow}
\usepackage{array}
\usepackage[caption=false,font=normalsize,labelfont=sf,textfont=sf]{subfig}
\usepackage{textcomp}
\usepackage{stfloats}
\usepackage{url}
\usepackage{verbatim}
\usepackage{pifont}
\usepackage{graphicx}
\usepackage{booktabs}
\usepackage{xcolor}   
\usepackage{colortbl} % colors
\usepackage{graphicx}
\usepackage{cite}
\usepackage[most]{tcolorbox}
\usepackage{makecell}
\usepackage{arydshln}
\newcolumntype{M}[1]{>{\centering\arraybackslash}m{#1}} % 水平居中 + 垂直居中（需要 array + m）
\definecolor{darkgreen}{rgb}{0.0, 0.5, 0.0}
\definecolor{lightgreen}{RGB}{240,255,250}
\begin{document}

\title{SpaceEra++: A Unified Framework Towards 3D Spatial Reasoning in Video}

\author{Weili Guan,~\IEEEmembership{Member,~IEEE}, Haoyu~Zhang, Meng Liu,~\IEEEmembership{Member,~IEEE}, Qianlong~Xiang, Yaowei Wang,~\IEEEmembership{Member,~IEEE}, Liqiang Nie,~\IEEEmembership{Senior Member,~IEEE}% <-this % stops a space

\IEEEcompsocitemizethanks{
\IEEEcompsocthanksitem Weili Guan is with the School of Information Science and Technology, Harbin Institute of Technology (Shenzhen), Shenzhen 518055, China and the Shenzhen Loop Area Institute, Shenzhen 518045, China (e-mail: honeyguan@gmail.com).
\IEEEcompsocthanksitem Haoyu Zhang and Yaowei Wang are with the School of Computer Science and Technology, Harbin Institute of Technology (Shenzhen), Shenzhen 518055, China and Pengcheng Laboratory, Shenzhen 518000, China (e-mail: zhang.hy.2019@gmail.com; wangyw@pcl.ac.cn).
\IEEEcompsocthanksitem Meng Liu is with the School of Computer Science and Technology, Shandong Jianzhu University, Jinan 250101, China and Zhongguancun Academy, Beijing 100190, China (e-mail: mengliu.sdu@gmail.com).
\IEEEcompsocthanksitem Qianlong Xiang is with the School of Computer Science and Technology, Harbin Institute of Technology (Shenzhen), Shenzhen 518055, China, the City University of Hong Kong, Hong Kong SAR, China, and the Shenzhen Loop Area Institute, Shenzhen 518045, China (e-mail: xiangqianlongcs@gmail.com).
\IEEEcompsocthanksitem Liqiang Nie is with the School of Computer Science and Technology, Harbin Institute of Technology (Shenzhen), Shenzhen 518055, China (e-mail: nieliqiang@gmail.com).
\IEEEcompsocthanksitem Corresponding author: Meng Liu and Liqiang Nie.
% \IEEEcompsocthanksitem Yaowei Wang is with the School of Computer Science and Technology, Harbin Institute of Technology (Shenzhen), Shenzhen 518055, China and Pengcheng Laboratory, Shenzhen 518000, China (e-mail: ).
}% <-this % stops an unwanted space
% \thanks{Haoyu Zhang is with the School of Computer Science and Technology, Shandong University, Qingdao 266000, China (e-mail: zhang.hy.2019@gmail.com).}
% \thanks{Meng Liu is with the School of Computer Science and Technology, Shandong Jianzhu University, Jinan 250101, China (e-mail: mengliu.sdu@gmail.com).}
% \thanks{Liqiang Nie is a professor with Harbin Institute of Technology (Shenzhen).
% (e-mail: nieliqiang@gmail.com).}
}

\markboth{Journal of \LaTeX\ Class Files,~Vol.~14, No.~8, August~2021}%
{Shell \MakeLowercase{\textit{et al.}}: A Sample Article Using IEEEtran.cls for IEEE Journals}


\maketitle

\begin{abstract}
Visual-spatial understanding, defined as the ability to infer object relationships and scene layouts from visual inputs, is fundamental to downstream tasks such as robotic navigation and embodied interaction.
{
However, pre-trained vision-language models (VLMs) remain constrained by spatial uncertainty stemming from inherently 2D observations and by the scarcity of data for 3D spatial understanding. To address these limitations, we proposed a novel framework, SpaceEra, in the NeurIPS 2025 Spotlight paper. Although it achieved significant performance gains, we further observed that its effectiveness is hindered by insufficient input from scanning videos and weak reasoning constraints. To tackle these newly emerged challenges, we extend the original framework into a comprehensive system, termed SpaceEra++, which spans data construction, model design, training optimization, and prompting inference. Specifically, to alleviate input insufficiency, we introduce ScenePick, a frame sampling strategy that balances spatial coverage with object semantics to produce compact yet comprehensive scene representations. In addition, to enhance spatial reasoning, we develop SpaceAlign, which enforces pairwise object constraints by jointly exploiting absolute coordinates and relative spatial relations, thereby aligning optimization with spatial accuracy. Extensive experiments across multiple benchmarks demonstrate consistent improvements over strong baselines, while ablation studies validate both the individual and joint contributions of each component, and further analyses provide guidance for future research.
Code and data are available at \url{https://github.com/iLearn-Lab/NeurIPS25-SpaceEra}.
% To overcome data scarcity, we construct ScanForgeQA, a scalable question–answering dataset automatically constructed from diverse 3D simulation scenes, enabling robust spatial commonsense acquisition. 
% To mitigate input insufficiency, we design ScenePick, which performs frame sampling that integrates spatial coverage and object semantics, yielding a comprehensive yet concise scene representation. 
% To strengthen reasoning constraints, we develop Space-based GRPO, which introduces pairwise object constraints that jointly leverage absolute coordinates and relative spatial relations, aligning optimization with spatial accuracy.
}
\end{abstract}


\section{Introduction}
\IEEEPARstart{V}{isual}-spatial understanding, the ability to infer spatial relationships and the layout of objects from visual input, is a core component of human perception~\cite{yang2024thinking,jin2024llava}. From a single image, human observers can intuitively estimate distances, relative sizes, and even infer occluded structures. As intelligent systems become increasingly embedded in real-world applications such as autonomous driving~\cite{tiandrivevlm,zhang2023attribute}, robotic navigation~\cite{driess2023palm}, and augmented reality~\cite{chandrasegaran2024hourvideo,zhang2025exo2ego}, it becomes crucial to endow models with similar spatial reasoning capabilities for robust perception and interaction.

Unfortunately, a single image is inherently limited in capturing the complexity of real-world 3D scenes, constraining its utility in practical scenarios~\cite{chen2024spatialvlm,cheng2025spatialrgpt,cai2024spatialbot}. To address this, point clouds have become a mainstream representation for 3D scene understanding due to their ability to encode rich geometric information~\cite{chen2024ll3da,hong20233d}. Yet, generating high-quality point clouds typically requires expensive sensors and incurs significant computational overhead, limiting scalability and accessibility. 

\begin{figure}
  \centering
    \includegraphics[width=\linewidth]{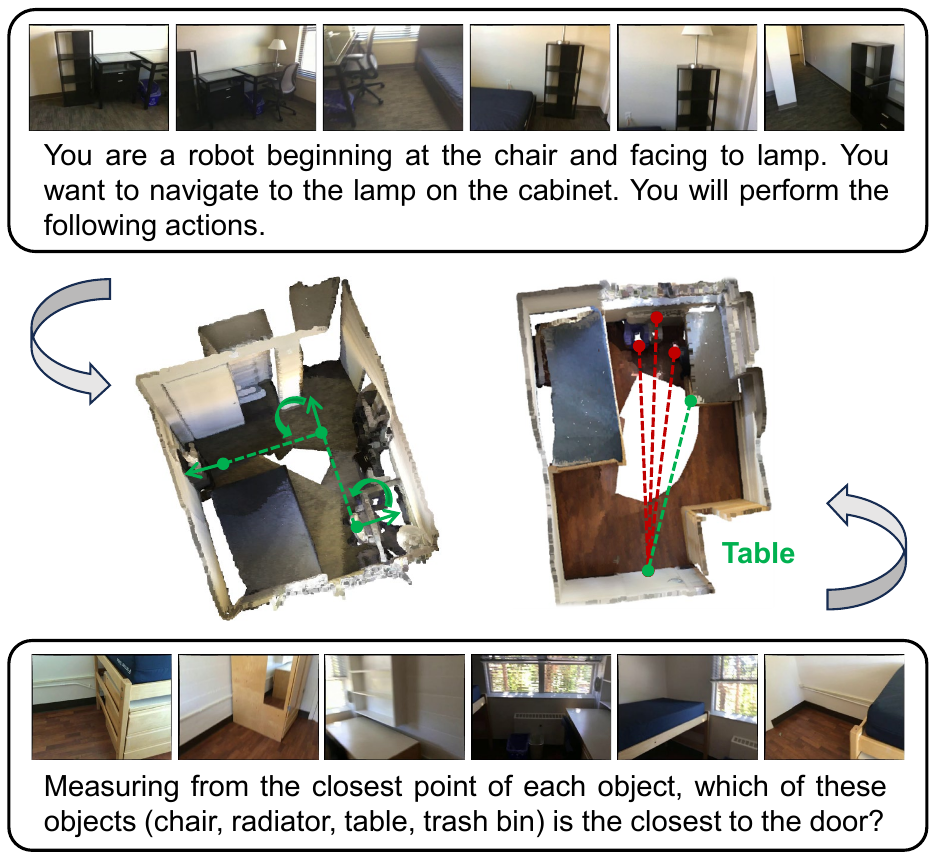}
    \caption{Two examples of spatial reasoning tasks: path planning (left) and relative distance (right).}
    \label{fig:task_exa}
\end{figure}

These limitations motivate the pursuit of vision-only solutions that operate on scanning videos or multi-view images of scenes, as shown in Fig.~\ref{fig:task_exa}. Such approaches offer a more human-like and scalable pathway to spatial understanding~\cite{qi2025gpt4scene}. However, performing 3D spatial reasoning from scanning videos presents two significant challenges:
% 数据、模型、训练策略
\ding{182} \textbf{Spatial Uncertainty}.
In the absence of explicit depth information, models must infer 3D structure from inherently limited 2D observations. This process is further complicated by occlusions, perspective distortions, and texture ambiguities, all of which introduce significant spatial uncertainty. 
% Effectively addressing this challenge demands multi-step logical reasoning across frames to reconstruct coherent spatial layouts. 
\ding{183} \textbf{Data Scarcity}.
Existing datasets for this task are limited in both scale and diversity, restricting the ability of vision-language models (VLMs) to acquire robust spatial knowledge and perceptual capabilities. 
% Moreover, these datasets involve scans of real-world scenes, which leads to poor scalability.
% This highlights the need for scalable and extensible data sources to support effective spatial reasoning in VLMs.
{
To address the two challenges above, we present an initial framework, SpaceEra, in our conference paper, which integrates the corresponding solutions. Specifically, we introduce \textbf{SpatialMind}, a structured Chain-of-Thought (CoT) prompting strategy that guides VLMs through step-by-step reasoning over spatial relationships to mitigate spatial uncertainty; and \textbf{ScanForgeQA}, a large-scale synthetic question-answering (QA) dataset constructed from diverse 3D simulation scenes via an automated generation pipeline to alleviate data scarcity.
% In our prior conference version, we proposed a framework that addressed above two primary challenges. Specifically, we introduce \textbf{SpatialMind}, a structured Chain-of-Thought (CoT) prompting strategy that guides VLMs through step-by-step reasoning over spatial relationships. And we present \textbf{ScanForgeQA}, a large-scale synthetic question-answering (QA) dataset constructed from diverse 3D simulation scenes using an automated generation pipeline. Through extensive experiments, we demonstrate that this framework can significantly improve the 3D spatial reasoning performance of VLMs under both zero-shot and fine-tuning settings.
\begin{figure}[t]
  \centering
    \includegraphics[width=\linewidth]{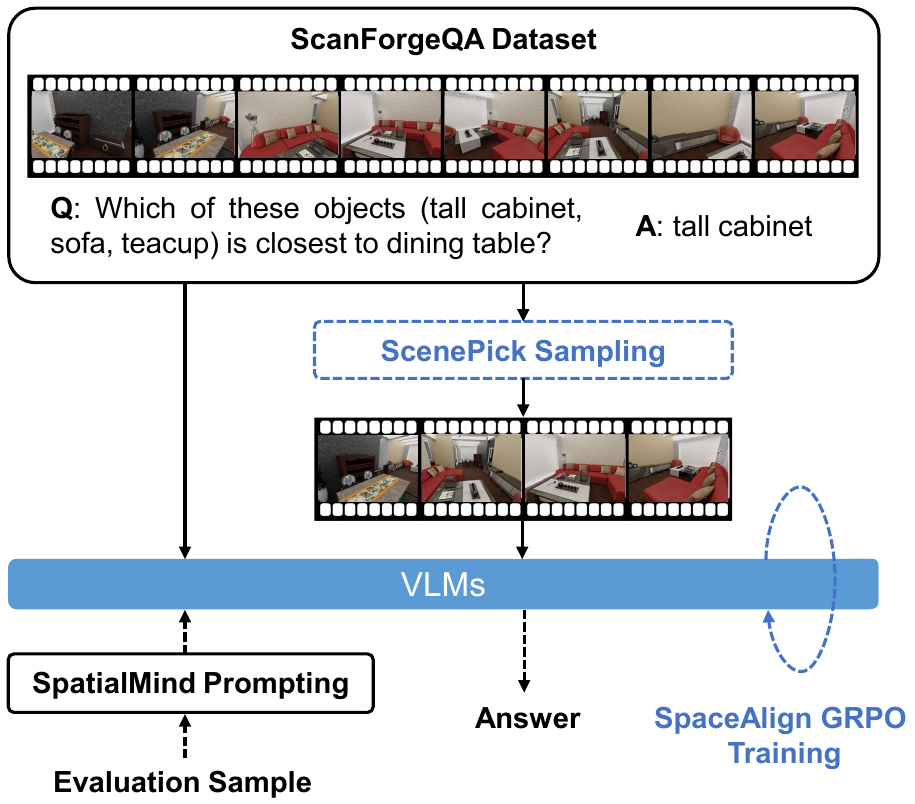}
    \caption{{Overview of the SpaceEra++ framework. The blue text denotes newly added content in this journal version. Solid black arrows indicate the training flow, while dashed black arrows represent the inference flow.}}
    \label{fig:extension}
\end{figure}

Although the above methods have achieved significant progress, several important challenges in model design and optimization remain insufficiently explored. To address this gap, in this work we further identify and systematically investigate the following issues:
% Although the above methods have achieved significant progress, the following challenges remain in model design and optimization:
\ding{184} \textbf{Input Insufficiency}.
Scanning videos of scenes often contain thousands of frames, yet existing VLMs can typically process only short sequences of 8 to 32 frames. The commonly used uniform sampling strategy inevitably discards important spatial cues and fine-grained scene details that are critical for spatial understanding. Therefore, it is essential to explore more effective frame sampling strategies that can better preserve and represent the semantic structure of the scene.
\ding{185} \textbf{Biased Reasoning}.
Recent studies have explored reinforcement learning (RL) approaches to improve reasoning capabilities. However, most of these methods focus mainly on achieving accurate alignment between predicted and ground-truth coordinates. This paradigm tends to overlook the relative spatial relationships among objects, which are often more important for understanding overall scene layouts. These observations underscore the need to jointly consider both absolute accuracy and relative relations.

% \begin{table}[t]
% \caption{A unified framework for zero-shot and fine-tuning settings.}
% \centering
% \renewcommand{\arraystretch}{1.2}
% {
% \begin{tabular}{lcc}
% \toprule
% \textbf{} & \textbf{Zero-shot} & \textbf{Fine-tune} \\
% \midrule
% \textbf{Data} & \textemdash & ScanForgeQA Dataset \\
% & \textemdash & $\downarrow$ \\
% \textbf{Model} & SpatialMind Prompting & ScenePick Sampling \\
% & \textemdash & $\downarrow$ \\
% \textbf{Training} & \textemdash & Space-based GRPO \\
% \bottomrule
% \end{tabular}\label{framework}
% }
% \end{table}

To address these newly identified challenges, we extend the original SpaceEra framework beyond its initial focus on data construction and inference guidance, and propose an enhanced architecture, \textbf{SpaceEra++}, for visual-spatial reasoning, as illustrated in Fig.~\ref{fig:extension}.
Specifically, we introduce \textbf{ScenePick}, a novel frame sampling strategy that improves the input stage by selecting informative and semantically representative frames from long scanning videos, thereby alleviating input insufficiency. Furthermore, we propose \textbf{SpaceAlign}, a reinforcement learning strategy that incorporates both absolute coordinate supervision and pairwise spatial constraints, enabling the model to capture relative object relationships and mitigate biased reasoning.
Together with the original SpaceEra design, these components form a unified framework that systematically addresses data, input, optimization, and inference in a coordinated manner.
We validate the proposed framework through extensive experiments across multiple benchmarks. The results not only demonstrate the effectiveness of each component, but also confirm that they provide complementary improvements when integrated into the overall framework, leading to substantial gains.

% To address these new challenges, we refine the existing framework and propose a more comprehensive architecture, SpaceEra++, to enhance 3D spatial reasoning of VLMs, as shown in Fig.~\ref{fig:extension}.
% Specifically, we propose \textbf{ScenePick}, a novel frame sampling strategy that balances scene coverage and object semantics, enabling the extraction of task-specific key frames to recover essential scene details. 
% Besides, considering the relative relationships among objects, we propose \textbf{SpaceAlign}, a reinforcement learning strategy with pairwise object constraints to integrate absolute coordinate accuracy and relative spatial relations, thereby enhancing the understanding of VLMs for spatial layouts.
% We have validated our overall framework through extensive experiments across multiple benchmarks. Results demonstrate the individual and combined effectiveness of our proposed extensions, and yield insights that may inspire future research on visual-spatial understanding.

A preliminary conference version of this work was presented as a Spotlight paper at NeurIPS 2025~\cite{zhang2025spatial}. The contributions of this journal version are summarized as follows:
\begin{itemize}
    \item We analyze the limitations of original schema and integrate the two newly proposed methods to form a unified technical framework for visual-spatial reasoning, spanning data, input, optimization, and inference.
    \item To mitigate insufficient visual input, we propose ScenePick, a frame sampling strategy that harmonizes spatial coverage and object semantics to ensure comprehensive scene understanding while preserving critical object details.
    \item To overcome biased reasoning optimization, we design SpaceAlign, a spatially constrained reinforcement learning algorithm that incorporates both absolute coordinates and relative relationships to enable more comprehensive spatial reasoning.
    \item Extensive experiments validate our overall framework, demonstrating both the standalone and complementary benefits of the proposed extensions, and providing detailed analysis of the results.
\end{itemize}

\section{Related Work}

\subsection{2D Image Spatial Understanding} 
This task focuses on modeling spatial relationships among objects within the 2D image. Most existing models are trained on 2D images paired with textual descriptions, which offer limited cues about 3D structure. Consequently, their capacity for spatial reasoning remains constrained. To mitigate this,  several approaches, such as SpatialVLM~\cite{chen2024spatialvlm}, SpatialRGPT~\cite{cheng2025spatialrgpt}, and SpatialBot~\cite{cai2024spatialbot}, have been proposed. These methods enhance the spatial understanding by fine-tuning models on datasets specifically designed for spatially grounded QA tasks. 
To enable more comprehensive evaluation, recent studies~\cite{du2024embspatial,zhang2024sphere} have introduced hierarchical benchmarks that assess models across varying levels of spatial reasoning complexity.
Parallel efforts have explored more explicit forms of spatial interaction~\cite{maspatialpin,yu2024rag}.
For example, point-based methods~\cite{yuanrobopoint,song2024robospatial} interpret spatial instructions by predicting specific target points. Building on this trend, SpatialCoT~\cite{liu2025spatialcot} proposes a two-stage strategy that aligns multimodal inputs with spatial coordinates and incorporates CoT reasoning to better address complex embodied tasks.
Despite these advancements, model performance often degrades in complex real-world 3D environments, highlighting the limitations of 2D-based approaches in representing complex 3D scenes.  
%their performance degrades significantly when transferred to real-world  environments. This decline is primarily attributed to the inherent limitations of independent 2D images in representing complex 3D scenes.

\begin{figure*}[t]
  \centering
  \includegraphics[width=0.9\linewidth]{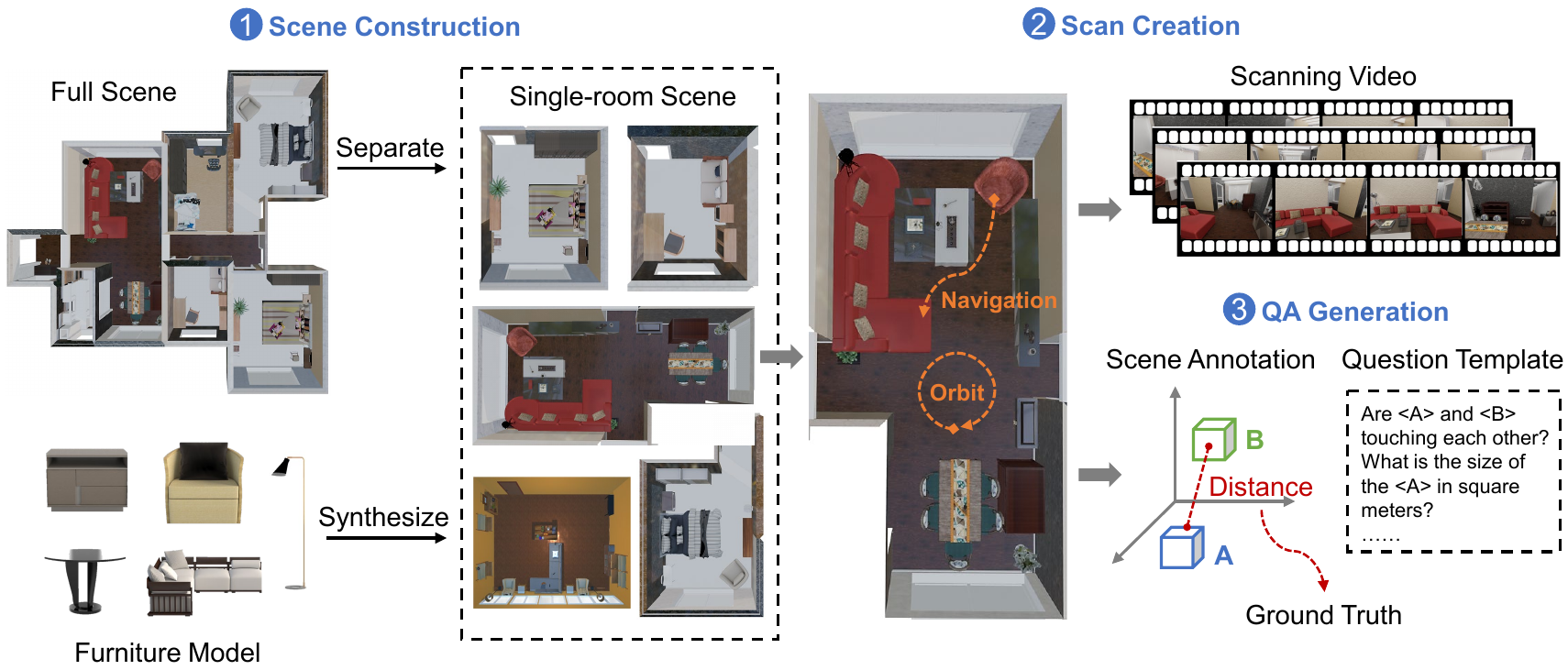}
  \caption{The pipeline of ScanForgeQA data construction, which consists of scene construction, scan creation, and QA generation.}
  \label{fig:data_gen}
  % \vspace{-2ex}
\end{figure*}

\subsection{3D Indoor Spatial Understanding}
This task focuses on enabling intelligent agents to identify object positions and infer their spatial relationships within enclosed environments, thereby supporting both object manipulation and interactive scene comprehension. Early 3D models are trained on standard indoor datasets~\cite{baruch1arkitscenes,chang2017matterport3d,dai2017scannet,deitke2022️,mao2022multiscan,ramakrishnan2habitat,yeshwanth2023scannet++} using point clouds to facilitate downstream tasks like 3D object detection and instance segmentation~\cite{vu2022softgroup,wu2024point,nguyen2024open3dis,rozenberszki2024unscene3d} and primarily focus on object-level geometry and appearance features~\cite{guo2023point,qi2024shapellm,liu2024lion,xu2024pointllm}. More recent work extends this focus to complex indoor scenes, emphasizing inter-object spatial relationships and holistic scene-level understanding. To address challenges such as geometric complexity and annotation sparsity, many of these models employ cross-modal strategies that combine point cloud data with auxiliary multi-view 2D images~\cite{chen2024ll3da,hong20233d,man2024lexicon3d}.
Inspired by the way humans perceive spatial layouts through vision alone, emerging research~\cite{yang2024thinking,liao2025improved,qi2025gpt4scene,zhu2025struct2d,ouyang2025spacer} has begun to explore purely vision-based approaches to 3D spatial understanding. These methods rely solely on visual inputs, such as scanning videos, without requiring explicit 3D priors like point clouds. This line of work offers a more practical and scalable alternative for real-world deployment. In this context, we further investigate whether purely vision-based inputs can provide a more effective solution for indoor scene understanding.

{
\subsection{Reinforcement Learning with Verifiable Reward}
Recent advances such as o1~\cite{jaech2024openai}, DeepSeek-R1~\cite{guo2025deepseek}, and Kimi K1.5~\cite{team2025kimi} have demonstrated that RL can substantially enhance the reasoning capabilities of large language models (LLMs). Among these efforts, the Group Relative Policy Optimization (GRPO) algorithm~\cite{shao2024deepseekmath} used in DeepSeek-R1 is particularly notable, as it highlights the strong potential of Reinforcement Learning with Verifiable Reward (RLVR) in equipping VLMs with advanced reasoning abilities. Building upon this foundation, subsequent studies have extended RLVR to boost multimodal reasoning performance. For example, LMM-R1~\cite{peng2025lmm} employs a two-stage rule-based RL pipeline to enable strong mathematical reasoning in 3B-scale models. To improve visual-spatial reasoning, several works have explored GRPO-based reinforcement learning. For example, \cite{liao2025improved} adopts simple answer-based rewards to show that GRPO alone can activate spatial reasoning. Spatial-MLLM~\cite{wu2025spatial} designs task-specific reward functions tailored to different categories of spatial reasoning, while SpaceR~\cite{ouyang2025spacer} introduces an absolute-coordinate estimation reward, enabling models to directly infer object positions and substantially enhancing spatial reasoning performance. However, these methods primarily focus on task correctness while overlooking the more essential aspect of object spatial layout in visual-spatial reasoning. In this work, we incorporate object coordinate accuracy and relative spatial relations as additional reward signals to enhance the model’s spatial imagination and layout understanding.

\section{Overall Framework}
As illustrated in Fig.~\ref{fig:extension}, we integrate the proposed extensions with the original framework to form a unified technical pipeline spanning data construction, model design, training optimization, and prompting inference. Specifically, using the constructed ScanForgeQA as training data, we take scene scanning videos together with QA pairs as input. The ScenePick frame sampling strategy is first applied to select key frames that jointly ensure broad spatial coverage and rich object semantics. These sampled frames, along with the corresponding QA pairs, are then fed into VLMs, where model parameters are optimized via the proposed SpaceAlign GRPO training strategy to enhance spatial reasoning capability. During inference, the trained model is guided by the SpatialMind prompting strategy to generate accurate predictions. We next provide a detailed description of each component in sequence.
}

\begin{table*}[t]
  \caption{Comparative analysis between ScanForgeQA and other existing 3D QA datasets.}
  \centering
{
\begin{tabular}{lcccccc}
\toprule
\textbf{Dataset} & \textbf{Data Source} & \textbf{Scene Number} & \textbf{Scene Format} & \textbf{Question Type} & \textbf{QA Number} & \textbf{Scalability} \\
\midrule
SPARTUN3D~\cite{zhang2024spartun3d} & 3RScan & 478 & Point Cloud & 4 & 133K & Difficult \\
MSQA~\cite{linghu2024multi} & ScanNet/3RScan/ARKitScenes & 1.7K & Point Cloud & 9 & 251K & Difficult \\
3D-LLM~\cite{hong20233d} & ScanNet/HM3D & 1.2K & Point Cloud & 9 & 300K & Difficult \\
\rowcolor{gray!20}ScanForgeQA & Simulation/Synthesis & 34K & Scan Video & 9 & 925K & Easy \\
\bottomrule
\end{tabular}
}
  \label{tab:data_comparison}
\end{table*}

\section{ScanForgeQA Dataset Construction}
The construction of the \textbf{ScanForgeQA} dataset involves a three-stage pipeline, illustrated in Fig.~\ref{fig:data_gen}. These stages are: \textbf{1) Scene Construction}, where single-room 3D environments are created; \textbf{2) Scan Creation}, in which egocentric videos are simulated by scanning through the constructed scenes; and \textbf{3) QA Generation}, where textual question-answering pairs are automatically generated based on object annotations and the spatial layout of each scene.

\subsection{Scene Construction}
To ensure diversity and richness in single-room scene collection, we adopt two parallel strategies:

% 可添加数据集示例
\subsubsection{\textbf{Separation}}
We modify existing scene datasets to leverage available resources effectively. Specifically, we utilize the 3D-FRONT dataset~\cite{fu20213d}, which contains 6,813 multi-room scenes furnished with diverse 3D objects and annotated with detailed layout semantics and high-quality textures. Since our focus is on single-room environments, we disassemble each multi-room scene into individual rooms. For each scene, we isolate and load one room at a time, along with its corresponding ceiling and walls, and save it as an independent instance. This disassembly process yields 44,427 single-room scenes. We further filter out uncommon room types (e.g., garage, auditorium) and those lacking sufficient object content (e.g., aisle, stairwell). The final dataset consists of 34,116 single-room scenes across six common categories: bedroom, kitchen, bathroom, living room, dining room, and storage room.
% And the type label of each room is also recorded according to the scene annotations. 

% \begin{figure}[t]
%   \centering
%   \includegraphics[width=\linewidth]{img/room_disv2.pdf}
%   \caption{Distribution of room types in the ScanForgeQA dataset.}
%   \label{fig:room_dis}
% \end{figure}

\subsubsection{\textbf{Synthesis}}
To introduce additional diversity and originality, we synthesize novel room layouts using a LLM-guided generation approach. Specifically, we adopt HoloDeck~\cite{yang2024holodeck}, a 3D generation framework that leverages LLMs to parse natural language prompts, retrieve matching assets from large-scale 3D object repositories such as Objaverse~\cite{deitke2023objaverse}, and optimize their spatial arrangement to form semantically meaningful scenes. To drive the generation process, we first use GPT-4o to create diverse textual descriptions for various room types. For example, a bedroom may be described as: \textit{``A bedroom with a bed, window, armchair, and wardrobe''}. We define eight room categories, including two additional types—office and store—and generate 20 distinct descriptions for each. These prompts are fed into HoloDeck to produce corresponding room layouts, with human verification to ensure spatial plausibility and realism. This synthesis process yields 160 additional single-room scenes.

% % 问题类型分布
% \begin{table*}[h]
% \caption{Data distribution.}
% \centering
% \begin{tabular}{ccccccccc}
% \toprule
%  & \makecell{Living\\Room} & \makecell{Kitchen} & \makecell{Bathroom} & \makecell{Bedroom} & \makecell{Office} & \makecell{Restaurant} & \makecell{Storage\\Room} & \makecell{Store} \\ 
% \midrule
% Number &  &  &  &  &  &  &  &  \\ 
% Scan &  &  &  &  &  &  &  &  \\
% QA &  &  &  &  &  &  &  &  \\
%     \bottomrule
% \end{tabular}
% \label{tab:example}
% \end{table*}
% 抖动

% \begin{figure}[t]
%   \centering
%     \includegraphics[width=\linewidth]{img/scanforgeqa_exa.pdf}
%     \caption{Examples of the proposed ScanForgeQA dataset.}
%     \label{fig:scanforgeqa_exa}
% \end{figure}

\subsection{Scan Creation}
To simulate egocentric scanning videos from the constructed single-room scenes, we implement a scanning procedure using the Unity engine. Each scene is scanned using two complementary strategies designed to emulate natural human visual exploration:

\subsubsection{\textbf{Orbit Scan}}
We define a circular trajectory centered in the room at a height of approximately 1.5 meters, corresponding to typical adult eye level. The circle’s diameter is set to two-thirds of the shorter side of the room. The camera is randomly initialized at a point on this path and moves along the circle either clockwise or counterclockwise. An image is captured every 5 degrees of rotation, resulting in 72 frames per orbit scan. This strategy provides a comprehensive 360-degree panoramic view of the scene.

\subsubsection{\textbf{Navigation Scan}}
To simulate movement through the environment, we label navigable ground regions based on object categories and generate a navigation mesh using the \textit{NavMesh Baking} API. We randomly select two objects as the navigation start and end points and compute the shortest path between them on the mesh. Among the candidate paths, the two longest are chosen for scanning to achieve a more complete coverage of the scene. For each path, the camera first performs a 360-degree rotation at the starting point, capturing an image every 12 degrees (30 images total). It then traverses the path toward the destination, during which 12 frames are uniformly sampled. Upon arrival, another 360-degree rotation is performed, again capturing 30 images. In total, 72 frames are recorded per path. Due to the limited size of indoor environments, rotational movement yields more visual variation than translation; hence, fewer frames are captured during motion.

% 答案生成示意图
\subsection{QA Generation}
To generate diverse supervised fine-tuning (SFT) data and enhance the 3D spatial reasoning capabilities of existing VLMs, we define three categories of question types: attribute estimation, spatial reasoning, and hypothesis analysis.  These categories encompass both quantitative and qualitative dimensions, and cover both open-set and closed-set scenarios. Below, we describe each category in detail, along with the methodology for deriving corresponding ground-truth answers.

\subsubsection{\textbf{Attribute Estimation}}
This type focuses on static properties of objects and scenes, such as \textit{object count} (``How many chairs are in the room?''), \textit{object size} (``What is the length of the longest side of the refrigerator in meters?''), \textit{room size} (``What is the size of this room in square meters?''), and \textit{room type} (``Based on the object layout, what is the most likely type of room (e.g., kitchen)?''). {Ground-truth answers for these questions are directly derived from 3D scene annotations and object metadata provided in the dataset.} 
% % 或者第一张图修改为答案判断示意图
% \begin{wrapfigure}{r}{0.5\textwidth}  % r=右侧图，l=左侧图；0.4宽度可调整
%     \centering
%     \includegraphics[width=0.48\textwidth]{img/format.pdf}
%     \caption{Effects of different scene expression.}
%     \label{fig:format_compare}
% \end{wrapfigure}

\subsubsection{\textbf{Spatial Reasoning}}
This category targets inter-object spatial relationships, requiring models to infer positional and geometric properties such as distance, orientation, and contact. Representative question types include: \textit{relative distance} (``Which of these objects (refrigerator, couch, ceiling light) is closest to the TV?''), \textit{absolute distance} (``What is the distance between the couch and the table in meters?''), \textit{relative direction} (``If I am standing by sofa and facing the table, which side is the trash can on?''), and \textit{contact relationship} (``Is there a gap between the bed and the headboard?''). For distance-related questions, we compute Euclidean distances between object centroids in the global 3D coordinate space. For contact relationships, object dimensions are also considered to determine physical adjacency. To resolve relative direction, we define an object’s front as the side oriented toward the room center. Angular sectors are divided clockwise into four directional categories: right (45°–135°), back (135°–225°), left (225°–315°), and front (315°–45°). For example, an object located at 80° relative to the reference point is classified as being on the right. 

\begin{figure*}[t]
  \centering
  \includegraphics[width=0.85\linewidth]{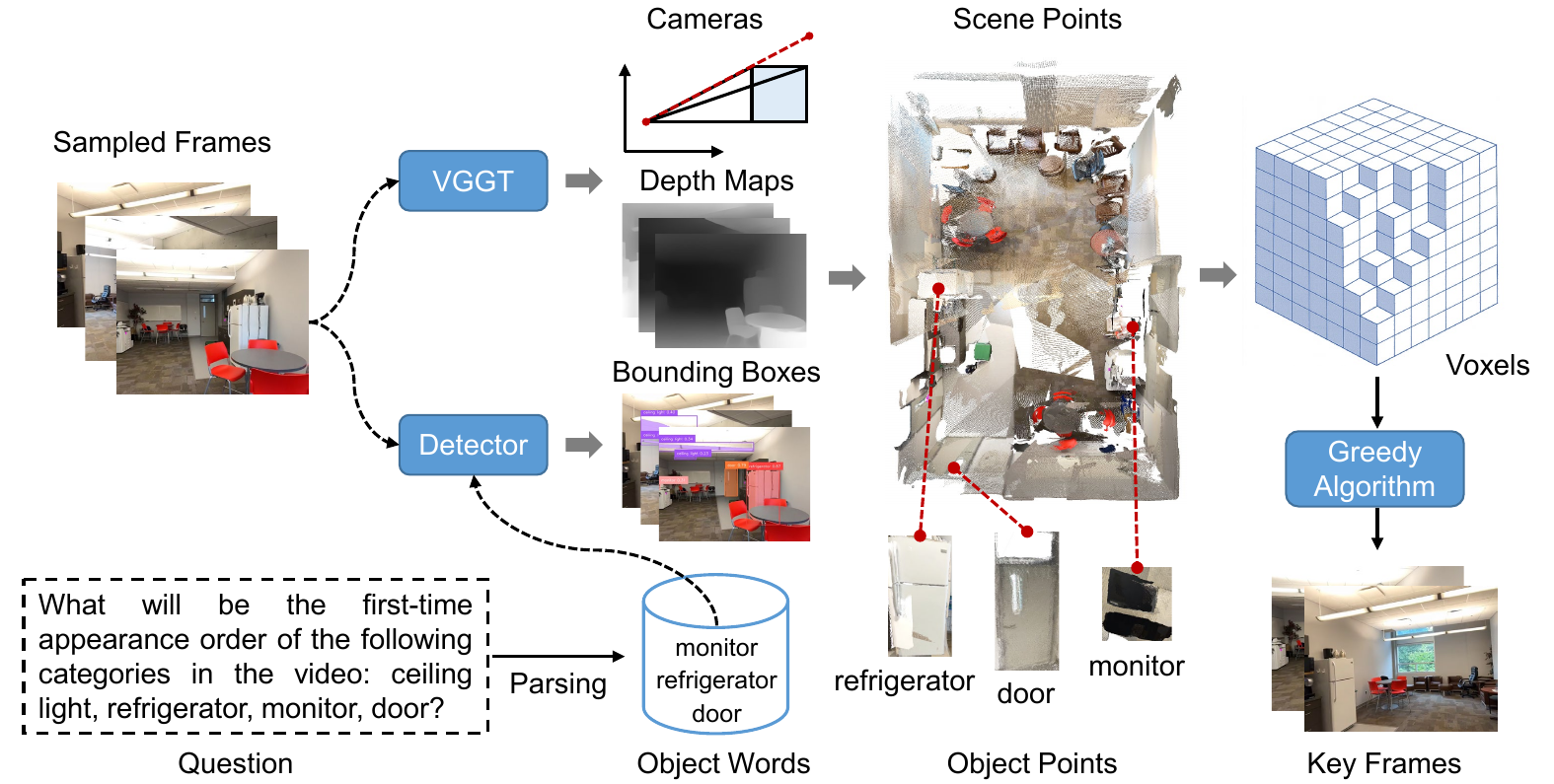}
  \caption{Illustration of our proposed ScenePick frame sampling strategy.}
  \label{fig:sampling}
\end{figure*}

\subsubsection{\textbf{Hypothesis Analysis}}
This category introduces conditional reasoning under hypothetical scenarios, often requiring geometric and commonsense inference.  A typical example is \textit{operation feasibility} (``Considering only object dimensions, is it feasible to place the television on the table?''). Feasibility is determined by comparing object dimensions. For stacking, the movable object’s length and width must be smaller than those of the supporting surface. For embedding (e.g., fitting an item into a drawer), the object’s height must also fall within the bounds of the specific container’s volume. 

{
% \subsection{Dataset Statistics}
A detailed comparison with existing datasets is provided in Table~\ref{tab:data_comparison}. The full ScanForgeQA dataset comprises 34,276 single-room scenes, 103K simulated video scans, and 925K QA pairs. 
}

% To assess which scene representation format is most interpretable for VLMs, we have conducted comparative experiments, as shown in Fig.~\ref{fig:format_compare}.

% \begin{algorithm}[t]
% \caption{ScenePick Frame Sampling}
% \label{alg:voxel_selection}
% {
% \begin{algorithmic}  % <-- [0] 关闭行号
% \REQUIRE Video sequence $\mathcal{V} = \{f_i\}_{i=1}^N$, target object set $\mathcal{O}$, number of selected frames $N_s$
% \ENSURE Selected frame subset $S$
% \STATE Uniformly sample $N_m$ frames $\{f_i^m\}_{i=1}^{N_m}$ from $\mathcal{V}$
% \STATE Decode depth maps $\{\mathbf{D}_i\}$ and camera poses $(\mathbf{E}_i, \mathbf{I}_i)$ using VGGT
% \STATE Reconstruct 3D point clouds $\mathcal{P}_i$ via reprojection
% \STATE Filter points with confidence $c(p) > 0.1$ to obtain $\mathcal{P}_{\text{valid}}$
% \STATE Compute voxel size $\Delta$ from the spatial extent of $\mathcal{P}_{\text{valid}}$
% \STATE Detect objects in $\mathcal{O}$ and assign weights to each point: $s(p) = 1$ if semantic, else $0.5$
% \STATE Map all valid points to voxels; compute voxel weights: $W(v) = \frac{1}{|\mathcal{P}_v|} \sum_{p \in \mathcal{P}_v} s(p)$
% \STATE Initialize $S \gets \emptyset$, $C \gets \emptyset$
% \REPEAT 
%   \FOR{each unselected frame $f_i^m$}
%     \STATE Compute gain: $G_i = \sum_{v \in V(f_i^m) \setminus C} W(v)$
%   \ENDFOR
%   \STATE Select $j = \arg\max_i G_i$
%   \STATE Update $S \gets S \cup \{j\}$, $C \gets C \cup V(f_j^m)$
% \UNTIL {$|S| \geq N_s$}
% \RETURN $S$
% \end{algorithmic}
% }
% \end{algorithm}

{
\section{ScenePick Frame Sampling}
Owing to constraints in context length and GPU memory, current VLMs are unable to process entire scene video sequences, and must instead work with only a small subset of frames. For example, in the VSI-Bench~\cite{yang2024thinking} evaluation, although each video sequence contains several thousand frames, the VLMs typically receive fewer than 64 frames as input. While uniform interval sampling is widely adopted and works well for standard video understanding tasks, spatial scanning videos present unique challenges as they capture 3D scene information. When it comes to spatial reasoning or 3D scene comprehension, uniform sampling often overlooks important spatial cues and critical scene details. To achieve a deeper and more accurate understanding of spatial scanning videos, it is necessary to develop frame sampling methods that prioritize the extraction of the most informative and representative aspects of the scene, rather than relying solely on evenly spaced selections in the time domain.

Building upon the capabilities of the feed-forward visual geometry foundation model~\cite{wang2025vggt}, we design a novel frame sampling strategy, ScenePick, which ensures comprehensive coverage of the scene’s spatial extent while selectively focusing on the key objects relevant to the question, as shown in Fig.~\ref{fig:sampling}.
Specifically, given a scene video $\mathcal{V} = \{f_i\}_{i=1}^{N}$, we first uniformly sample $N_m$ frames $\{f_i^m\}_{i=1}^{N_m}$ as a subset, where $N_m$ is determined by the available GPU memory. Our objective is to select $N_s$ frames $\{f_{i}^s\}_{i=1}^{N_s}$ from $\{f_i^m\}_{i=1}^{N_m}$ subset have most coverage of the underlying scene, where $N_s < N_m < N$. To achieve this, we leverage pretrained VGGT model~\cite{wang2025vggt} to decode a set of camera parameters and depth maps from the $N_m$ candidate frames:
\begin{equation}
    \{\mathbf{E}_i, \mathbf{I}_i, \mathbf{D}_i\}_{i=1}^{N_m} = \text{VGGT}(\{f_i^m\}_{i=1}^{N_m}),
\end{equation}
where $\mathbf{E}_i$ and $\mathbf{I}_i$ are extrinsic and intrinsic matrices of the camera, respectively, and $\mathbf{D}_i$ denotes the predicted depth map for each frame. Then we reconstruct 3D point clouds $\mathcal{P}_i$ for each frame by reprojecting depth values:
\begin{equation}
    \mathbf{x}_\mathrm{cam} = \mathbf{I}_i^{-1} \begin{bmatrix} u \\ v \\ 1 \end{bmatrix} \cdot \mathbf{D}_i, \qquad
    \mathcal{P}_i = \mathbf{E}_i^{-1} \begin{bmatrix} \mathbf{x}_\mathrm{cam} \\ 1 \end{bmatrix},
\end{equation}
where $(u, v)$ denotes the pixel coordinates. For each 3D point $p \in \mathcal{P}_i$, we also obtain a confidence score $c(p) \in [0, 1]$ from the depth head of VGGT. For subsequent processing, we aggregate all valid points across the sampled frames—those with confidence $c(p) > 0.1$—to define the spatial extent of the scene:
\begin{equation}
    \mathcal{P}_{\mathrm{valid}} = \bigcup_{i=1}^{N_m} \{ p \in \mathcal{P}_i\;|\; c(p) > 0.1 \}.
\end{equation}

We then discretize the 3D bounding box covering all valid points into voxels. To account for variations in scale in VGGT outputs, we set the voxel size $\Delta$ adaptively as a fraction of the bounding box’s dimension:
\begin{equation}
    \Delta = \frac{1}{\lambda} \cdot (\max(\mathcal{P}_{\mathrm{valid}}) - \min(\mathcal{P}_{\mathrm{valid}})),
    \label{eq:voxel_size}
\end{equation}
where $\lambda$ is a tunable hyperparameter. For each candidate frame, we then compute its voxel coverage $V(f_i^m)$ by mapping its valid 3D points into the voxel grid:
\begin{equation}
    V(f_i^m) = \left\{ \left\lfloor \frac{p - \min(\mathcal{P}_{\mathrm{valid}})}{\Delta} \right\rfloor \;\middle|\; p \in \mathcal{P}_i \cap \mathcal{P}_{\mathrm{valid}} \right\}.
    \label{eq:voxel_coverage}
\end{equation}

However, relying solely on spatial voxel coverage for frame selection is not sufficient. Different questions target different objects in the scene, which means the key frames required for reasoning can vary accordingly. Therefore, frame selection should also account for the semantic importance of the relevant objects. Specifically, we first parse the input natural language question to extract a set of target object categories (e.g., desk, chair, and sofa) relevant to the scene $\mathcal{O}=\{o_j\}$. For each candidate frame \(f^m_i\), we apply an object grounding model (e.g., Grounded SAM 2) over the set of target object categories \(\mathcal{O}\), obtaining a set of bounding boxes \(\mathcal{B}_i^{o_j}\) for each category \(o_j \in \mathcal{O}\).
For every pixel coordinate \((u,v)\) in frame  \(f^m_i\) and its corresponding 3D point \(p \in \mathcal{P}_i\), we define the semantic indicator function:
\begin{equation}
    s(p) \;=\;
    \begin{cases}
      1, & \exists\,o_j \in \mathcal{O}\;\text{s.t.}\;(u,v)\in \mathcal{B}_i^{o_j},\\
      0.5, & \text{otherwise}.
    \end{cases}
\end{equation}

We discretize the scene into a voxel grid. For each voxel $v \in V(f_i^m)$, let $\mathcal{P}_v$ denote the set of 3D points inside it. The weight of voxel $v$ is computed as the average weight of all its contained points:

\begin{equation}
W(v) = \frac{1}{|\mathcal{P}_v|} \sum_{p \in \mathcal{P}_v} s(p).
\end{equation}

Finally, we formulate the frame selection task as a classic maximum coverage problem. The objective is to select a subset $S$, such that the union of their voxel coverages is maximized:
\begin{equation}
\max_{S \subseteq \{1, \ldots, N_m\},\; |S| = N_s}
 \sum_{v \in \bigcup_{i \in S} V(f_i^m)} W(v).
\end{equation}

We adopt a greedy strategy, where at each step we select the frame that maximizes the gain, defined as the number of newly covered spatial voxels. This greedy selection process is repeated until $N_s$ frames $\{f_i^s\}_{i=1}^{N_s}$ are chosen. The resulting set of frames not only maximizes the spatial coverage of the scene but also ensures adequate semantic attention to objects relevant to the question.
}

\begin{algorithm}[t]
\caption{ScenePick Frame Sampling}
\label{alg:voxel_selection}
{
\begin{algorithmic}[1]
\REQUIRE Video $\mathcal{V}=\{f_i\}_{i=1}^{N}$, target object set $\mathcal{O}$, number of selected frames $N_s$
\ENSURE Selected frame subset $S$
\STATE Uniformly sample $N_m$ frames $\{f_i^m\}_{i=1}^{N_m}$ from $\mathcal{V}$
\STATE Use VGGT to decode depth maps $\{\mathbf{D}_i\}$ and camera poses $(\mathbf{E}_i,\mathbf{I}_i)$ for each $f_i^m$
\STATE Reproject depth to obtain 3D points $\mathcal{P}_i$ and filter points by confidence
\STATE Aggregate all valid points to form $\mathcal{P}_{\text{valid}}$ and compute voxel size $\Delta$
\STATE Detect object instances in $\mathcal{O}$ for each frame
\STATE Assign semantic weight $s(p)$ for each valid 3D point
\STATE Voxelize points of each frame to obtain voxel set $V(f_i^m)$ and per-voxel weight $W(v)=\frac{1}{|\mathcal{P}_v|}\sum_{p\in\mathcal{P}_v}s(p)$
\STATE Initialize selected set $S\gets\emptyset$ and covered voxel set $C\gets\emptyset$
\REPEAT
  \FOR{each unselected frame $f_i^m$}
    \STATE Compute marginal gain $G_i=\sum_{v\in V(f_i^m)\setminus C} W(v)$
  \ENDFOR
  \STATE Select $j=\arg\max_i G_i$
  \STATE Update $S\gets S\cup\{j\}$, $C\gets C\cup V(f_j^m)$
\UNTIL{$|S| = N_s$}
\RETURN $S$
\end{algorithmic}
}
\end{algorithm}

{
\section{Training}
Based on the above constructed ScanForgeQA dataset, we explore two training strategies: SFT and RLVR.
\subsection{Supervised Fine-tuning}
Leveraging the constructed ScanForgeQA dataset, we fine-tune several state-of-the-art models (e.g., Qwen2.5-VL and InternVL2) by applying the standard cross-entropy loss $\mathcal{L}_{\mathrm{CE}}$ between the predictions and the ground-truth annotations:
\begin{equation}
    \mathcal{L}_{\mathrm{CE}}(\theta) = -\sum_i \log P_{\theta}\left(o^{(i)} \,\middle|\, o^{(1:i-1)}, q, \{f_j\}_{j=1}^{N_s} \right),
\end{equation}
where $\{f_j\}_{j=1}^{N_s}$ denotes input video frames, $q$ denotes the prompt and question, $o^{(i)}$ represents the $i$-th token in the ground-truth answer, and $o^{(1:i-1)}$ denotes the preceding answer tokens.

\subsection{Reinforcement Learning with Verifiable Reward}
To enhance visual-spatial reasoning in VLMs, we propose a reinforcement learning framework, SpaceAlign, built upon GRPO \cite{guo2025deepseek}. SpaceAlign introduces verifiable reward functions tailored to diverse QA types and a novel space imagination mechanism to guide spatial reasoning.
\subsubsection{\textbf{Verifiable Reward Function}}
To supervise model outputs across multiple QA types, including multi-choice and numerical, we design a set of verifiable reward functions that assess either response
format or correctness based on task-specific criteria.

\textbf{Format Reward}. To ensure the model responses adhere to a predefined structure, we define a format reward 
$R_{\mathrm{format}}$ based on whether the model wraps its reasoning process and answer within 
\texttt{<think>...</think>} and \texttt{<answer>...</answer>} tags, respectively. Specifically, $R_{\mathrm{format}}(\hat{y}) = 1$ if $\hat{y}$ matches the required format, and $R_{\mathrm{format}}(\hat{y}) = 0$ otherwise, where $\hat{y}$ is the model’s response.

% \begin{equation}
% R_{\text{format}}(\hat{y}) =
% \begin{cases}
% 1, & \text{if } \hat{y} \text{ matches format}, \\
% 0, & \text{otherwise.}
% \end{cases}
% \end{equation}

\textbf{Task Reward}. Different reward calculation strategies are employed based on the specific types of answers. For multi-choice QA, the reward $R_{\mathrm{mc}}$ is binary, based on exact match with the ground truth: $R_{\mathrm{mc}}(\hat{y}, y) = 1$ if $\hat{y} = y$, and $R_{\mathrm{mc}}(\hat{y}, y) = 0$ otherwise, 
% \begin{equation}
% R_{\text{mc}}(\hat{y}, y) =
% \begin{cases}
% 1, & \text{if } \hat{y} = y, \\
% 0, & \text{otherwise,}
% \end{cases}
% \end{equation}
where $y$ is the ground truth.
To assess numerical values, we compute relative accuracy across varying confidence thresholds 
$\theta \in \{0.5, 0.55, \ldots, 0.95\}$.
The numerical reward $R_{\mathrm{num}}$ is defined as:
\begin{equation}
R_{\mathrm{num}}(\hat{y}, y)
= \frac{1}{N} \sum_{i=1}^{N}
\mathbb{I}\left(
\frac{|\hat{y} - y|}{y} \leq 1 - \theta_i
\right),
\end{equation}
where $N$ is the number of confidence thresholds.

\subsubsection{\textbf{Space Imagination Reward}} 
Since existing RLVR frameworks such as GRPO lack explicit reward signals for visual-spatial reasoning, we introduce a space imagination mechanism. Specifically, the model is guided to generate an $M \times M$ grid map representing object distributions within the scene, which facilitates downstream reasoning and leads to more reliable answers.

\textbf{Absolute Coordinate Reward}. 
To quantitatively assess the quality of the generated spatial map, 
we propose a novel reward function $R_{\mathrm{abs}}$ that measures the spatial accuracy 
between predicted and ground truth object positions. 
We first compute the distance between each predicted object and its ground truth counterpart as:
\begin{equation}
    d_i= 
\frac{\sqrt{(x_{p,i} - x_{g,i})^2 + (y_{p,i} - y_{g,i})^2}}{\sqrt{M^2 + M^2}},
\end{equation}
where $(x_{p,i}, y_{p,i})$ and $(x_{g,i}, y_{g,i})$ denote the coordinates of the $i$-th predicted and ground truth objects, respectively,
and $M$ is the grid map size.
The reward $R_{\mathrm{abs}}$ is then computed as the weighted mean of spatial accuracies over all object categories:
\begin{equation}
R_{\mathrm{abs}} =
\sum_{i=1}^{k}
\left(
\frac{n_i}{\sum_{j=1}^{k} n_j}
\times
\left(
1 - d_i
\right)
\right),
\end{equation}
where $k$ is the total number of object categories, and $n_i$ denotes the number of objects in the $i$-th category. 

\begin{figure*}
  \centering
  \includegraphics[width=0.9\linewidth]{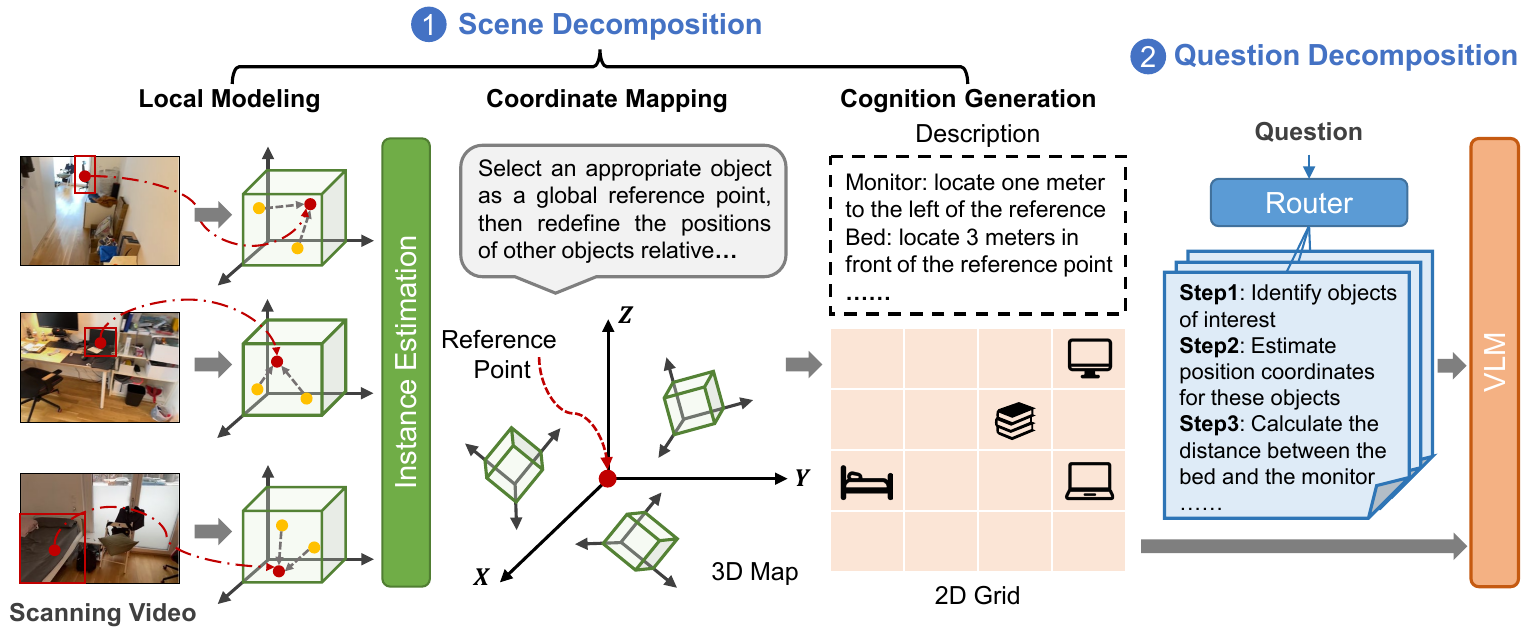}
  \caption{Illustration of our SpatailMind prompting strategy, which consists of two main steps: scene decomposition and question decomposition.}
  \label{fig:cot}
  % \vspace{-2ex}
\end{figure*}

\begin{table*}[ht]
  \caption{Performance comparison on VSI-Bench. \textcolor{red}{\raisebox{0.2ex}{$\dagger$}} indicates results on VSI-Bench (tiny) set. \textcolor{green!15!white}{\rule{2em}{1em}} indicates the current best results, while \textcolor{lightgreen}{\rule{2em}{1em}} represents the best results from the conference version.}

  \centering
  \begin{tabular}{lccccccccccc}
    \toprule
    \textbf{Method} & \textbf{Backbone}&
    \makecell{\textbf{Obj.} \\ \textbf{Count}} & 
    \makecell{\textbf{Abs.} \\ \textbf{Dist.}} & 
    \makecell{\textbf{Obj.} \\ \textbf{Size}} & 
    \makecell{\textbf{Room} \\ \textbf{Size}} & 
    \makecell{\textbf{Rel.} \\ \textbf{Dist.}} & 
    \makecell{\textbf{Rel.} \\ \textbf{Dir.}} & 
    \makecell{\textbf{Route} \\ \textbf{Plan}} & 
    \makecell{\textbf{Appr.} \\ \textbf{Order}} & 
    \textbf{Avg}&
    $\Delta$\\

    % \textbf{Method} & \textbf{Obj. Count} & \textbf{Abs. Dist.} & \textbf{Obj. Size} & \textbf{Room Size} & \textbf{Rel. Dist.} & \textbf{Rel. Dir.}$^{\star}$ & \textbf{Room Plan} & \textbf{Appr. Order} & \textbf{Avg.}\\
    \midrule
    \rowcolor{gray!10}\multicolumn{12}{c}{Close-source Models}\\
    \midrule
     Human Level\textcolor{red}{\raisebox{0.2ex}{$\dagger$}}&& 94.3 & 47.0 & 60.4 & 45.9& 94.7& 95.8& 95.8 &100.0& 79.2&-\\
     Gemini-2.0 Flash\textcolor{red}{\raisebox{0.2ex}{$\dagger$}}&& 52.4&30.6&66.7&31.8&56.0&46.3&24.5&55.1&45.4&-\\
     Gemini-1.5 Pro\textcolor{red}{\raisebox{0.2ex}{$\dagger$}}&\multirow{4}{*}{Gemini-1.5 Pro}& 49.6& 28.8& 58.6& 49.4& 46.0& 48.1&  42.0&  68.0& 48.8&-\\
     Gemini-1.5 Pro &&  56.2 & 30.9 & 64.1& 43.6& 51.3& 46.3& 36.0& 34.6& 45.4&-\\
     SpaceEra & 
& \cellcolor{lightgreen}63.9
& \cellcolor{lightgreen}51.8
& \cellcolor{lightgreen}70.2
& \cellcolor{lightgreen}47.3
& \cellcolor{lightgreen}56.3
& \cellcolor{lightgreen}45.9
& \cellcolor{lightgreen}42.6
& \cellcolor{lightgreen}44.3
& \cellcolor{lightgreen}52.8
& \cellcolor{lightgreen}\textbf{\color{darkgreen}{$\uparrow$ 7.4\%}} \\

     % \rowcolor{lightgreen}+SpatialMind&&63.9&51.8&70.2&47.3&56.3&45.9&42.6&44.3&52.8&\textbf{{\color{darkgreen} $\uparrow$ 7.4\%}}\\
     +ScenePick&&59.7&33.5&68.8&46.5&53.4&50.1&39.6&40.3&49.0&\textbf{{\color{darkgreen} $\uparrow$ 3.6\%}}\\
     GPT-4o&\multirow{2}{*}{GPT-4o}&46.2 &5.3 &43.8& 38.2 &37.0& 41.3& 31.5& 28.5&34.0&-\\
     SpaceEra & 
& \cellcolor{lightgreen}40.0
& \cellcolor{lightgreen}27.1
& \cellcolor{lightgreen}62.7
& \cellcolor{lightgreen}40.9
& \cellcolor{lightgreen}41.0
& \cellcolor{lightgreen}39.6
& \cellcolor{lightgreen}37.1
& \cellcolor{lightgreen}38.5
& \cellcolor{lightgreen}40.8
& \cellcolor{lightgreen}\textbf{\color{darkgreen}{$\uparrow$ 6.8\%}} \\

     % \rowcolor{lightgreen}SpaceEra&&40.0&27.1&62.7&40.9&41.0&39.6&37.1&38.5&40.8&\textbf{{\color{darkgreen} $\uparrow$ 6.8\%}}\\
     % +ScenePick\\
     \midrule
    \rowcolor{gray!10}\multicolumn{12}{c}{Open-source Models}\\
    \midrule
     % DeepSeek& \\
     % +SpatialMind&\\
    Struct2D&\multirow{2}{*}{Qwen2.5-VL-3B}&46.0&34.7&56.4&42.6&35.1&44.9&33.5&-&41.9&-\\
    Spatial-MLLM&&65.3&34.8&63.1&45.1&41.3&46.2&33.5&46.3&48.4&-\\
    SpaceVista&Qwen2.5-VL-7B&62.9&36.0&58.1&42.0&44.2&49.7&38.9&56.3&48.6&-\\
    \midrule
     InternVL2-8B &\multirow{5}{*}{InternVL2-8B}& 23.1 &28.7& 48.2 &39.8& 36.7 &30.7& 29.9 &39.6& 34.6&-\\
     % +SpatialMind&35.8&28.9&49.7&44.4&37.2&34.8&35.1&45.5&38.9&\textbf{{\color{darkgreen} $\uparrow$ 4.3\%}}\\
     
     % +ScanForgeQA&45.3&33.4&54.8&45.0&41.1&36.1&33.4&43.0&41.5&\textbf{{\color{darkgreen} $\uparrow$ 6.9\%}}\\
     SpaceEra & 
        & \cellcolor{lightgreen}47.0
        & \cellcolor{lightgreen}32.8
        & \cellcolor{lightgreen}53.2
        & \cellcolor{lightgreen}46.6
        & \cellcolor{lightgreen}39.8
        & \cellcolor{lightgreen}36.8
        & \cellcolor{lightgreen}37.9
        & \cellcolor{lightgreen}47.5
        & \cellcolor{lightgreen}42.7
        & \cellcolor{lightgreen}\textbf{\color{darkgreen}{$\uparrow$ 8.1\%}} \\
        % SpaceEra&&\rowcolor{lightgreen}47.0&32.8&53.2&46.6&39.8&36.8&37.9&47.5&42.7&\textbf{{\color{darkgreen} $\uparrow$ 8.1\%}}\\
     +ScenePick&&47.8&34.1&54.3&49.3&42.8&39.1&39.8&48.2&44.4&\textbf{{\color{darkgreen} $\uparrow$ 1.7\%}}\\
     +SpaceAlign&&53.3&37.6&55.0&47.4&42.5&41.0&41.7&49.8&46.0&\textbf{{\color{darkgreen} $\uparrow$ 3.3\%}}\\
          SpaceEra++ & 
        & \cellcolor{green!15!white}54.8
        & \cellcolor{green!15!white}38.7
        & \cellcolor{green!15!white}55.2
        & \cellcolor{green!15!white}47.9
        & \cellcolor{green!15!white}43.1
        & \cellcolor{green!15!white}41.1
        & \cellcolor{green!15!white}42.2
        & \cellcolor{green!15!white}50.4
        & \cellcolor{green!15!white}46.7
        & \cellcolor{green!15!white}\textbf{\color{darkgreen}{$\uparrow$ 4.0\%}} \\
% \rowcolor{green!15!white}SpaceEra++&&54.8&38.7&55.2&47.9&43.1&41.1&42.2&50.4&46.7&\textbf{{\color{darkgreen} $\uparrow$ 4.0\%}}\\
          \cdashline{1-12}[1pt/1pt]
     InternVL2-40B &\multirow{5}{*}{InternVL2-40B}& 34.9& 26.9& 46.5& 31.8 &42.1& 32.2& 34.0 &39.6& 36.0&-\\
     SpaceEra & 
& \cellcolor{lightgreen}52.2
& \cellcolor{lightgreen}30.5
& \cellcolor{lightgreen}54.4
& \cellcolor{lightgreen}41.0
& \cellcolor{lightgreen}50.5
& \cellcolor{lightgreen}37.0
& \cellcolor{lightgreen}40.2
& \cellcolor{lightgreen}50.3
& \cellcolor{lightgreen}44.5
& \cellcolor{lightgreen}\textbf{\color{darkgreen}{$\uparrow$ 8.5\%}} \\
    % \rowcolor{lightgreen}SpaceEra&&52.2&30.5&54.4&41.0&50.5&37.0&40.2&50.3&44.5&\textbf{{\color{darkgreen} $\uparrow$ 8.5\%}}\\

     % +SpatialMind&36.4&30.0&49.1&41.8&43.8&36.1&35.6&50.0&40.4&\textbf{{\color{darkgreen} $\uparrow$ 4.4\%}}\\
     % +ScanForgeQA&51.0&29.2&52.7&38.1&47.2&36.4&35.9&47.6&42.3&\textbf{{\color{darkgreen} $\uparrow$ 6.3\%}}\\
     +ScenePick&&54.0&32.7&56.0&44.4&53.7&37.7&41.1&52.5&46.5&\textbf{{\color{darkgreen} $\uparrow$ 2.0\%}}\\
      +SpaceAlign&&57.9&32.3&58.3&45.0&56.4&42.4&44.6&52.7&48.7&\textbf{{\color{darkgreen} $\uparrow$ 4.2\%}}\\
      SpaceEra++ & 
& \cellcolor{green!15!white}59.2
& \cellcolor{green!15!white}33.2
& \cellcolor{green!15!white}58.9
& \cellcolor{green!15!white}46.5
& \cellcolor{green!15!white}58.1
& \cellcolor{green!15!white}42.8
& \cellcolor{green!15!white}45.1
& \cellcolor{green!15!white}53.2
& \cellcolor{green!15!white}49.6
& \cellcolor{green!15!white}\textbf{\color{darkgreen}{$\uparrow$ 5.1\%}} \\

          % \rowcolor{green!15!white}SpaceEra++&&59.2&33.2&58.9&46.5&58.1&42.8&45.1&53.2&49.6&\textbf{{\color{darkgreen} $\uparrow$ 5.1\%}}\\
    \midrule

    Qwen2.5-VL-7B&\multirow{5}{*}{Qwen2.5-VL-7B}&40.3&22.2&50.1&38.9&38.0&40.7&31.4&35.9&37.2&-\\
     SpaceEra & 
& \cellcolor{lightgreen}55.0
& \cellcolor{lightgreen}29.5
& \cellcolor{lightgreen}57.3
& \cellcolor{lightgreen}44.0
& \cellcolor{lightgreen}43.5
& \cellcolor{lightgreen}44.3
& \cellcolor{lightgreen}38.3
& \cellcolor{lightgreen}39.2
& \cellcolor{lightgreen}43.9
& \cellcolor{lightgreen}\textbf{\color{darkgreen}{$\uparrow$ 6.7\%}} \\
% \rowcolor{lightgreen}SpaceEra&&55.0&29.5&57.3&44.0&43.5&44.3&38.3&39.2&43.9&\textbf{{\color{darkgreen} $\uparrow$ 6.7\%}}\\
          +ScenePick&&57.3&30.2&58.5&45.6&45.3&45.4&39.0&39.9&45.2&\textbf{{\color{darkgreen} $\uparrow$ 1.3\%}}\\
     +SpaceAlign&&65.1&31.7&60.1&49.1&49.6&51.0&40.1&42.5&48.7&\textbf{{\color{darkgreen} $\uparrow$ 4.8\%}}\\
     SpaceEra++ & 
& \cellcolor{green!15!white}66.7
& \cellcolor{green!15!white}32.6
& \cellcolor{green!15!white}62.0
& \cellcolor{green!15!white}49.7
& \cellcolor{green!15!white}52.3
& \cellcolor{green!15!white}52.0
& \cellcolor{green!15!white}43.1
& \cellcolor{green!15!white}44.3
& \cellcolor{green!15!white}50.3
& \cellcolor{green!15!white}\textbf{\color{darkgreen}{$\uparrow$ 6.4\%}} \\

     % \rowcolor{green!15!white}SpaceEra++&&58.6&31.4&61.2&47.6&46.4&47.7&39.3&44.5&47.1&\textbf{{\color{darkgreen} $\uparrow$ 3.2\%}}\\
     \cdashline{1-12}[1pt/1pt]
        Qwen2.5-VL-72B&\multirow{3}{*}{Qwen2.5-VL-72B}&37.9& 28.6&57.4&49.8&45.5&38.4&20.6&35.4&39.2&-\\
        SpaceEra & 
& \cellcolor{lightgreen}48.6
& \cellcolor{lightgreen}34.4
& \cellcolor{lightgreen}68.9
& \cellcolor{lightgreen}54.7
& \cellcolor{lightgreen}53.4
& \cellcolor{lightgreen}43.9
& \cellcolor{lightgreen}30.1
& \cellcolor{lightgreen}42.7
& \cellcolor{lightgreen}47.1
& \cellcolor{lightgreen}\textbf{\color{darkgreen}{$\uparrow$ 7.9\%}} \\
     % \rowcolor{lightgreen}SpaceEra&&48.6&34.4&68.9&54.7&53.4&43.9&30.1&42.7&47.1&\textbf{{\color{darkgreen} $\uparrow$ 7.9\%}}\\
     +ScenePick & 
& \cellcolor{green!15!white}50.7
& \cellcolor{green!15!white}36.4
& \cellcolor{green!15!white}74.5
& \cellcolor{green!15!white}57.7
& \cellcolor{green!15!white}54.8
& \cellcolor{green!15!white}45.2
& \cellcolor{green!15!white}34.2
& \cellcolor{green!15!white}46.4
& \cellcolor{green!15!white}50.0
& \cellcolor{green!15!white}\textbf{\color{darkgreen}{$\uparrow$ 2.9\%}} \\

          % \rowcolor{green!15!white}+ScenePick&&50.7&36.4&74.5&57.7&54.8&45.2&34.2&46.4&50.0&\textbf{{\color{darkgreen} $\uparrow$ 2.9\%}}\\
        \cdashline{1-12}[1pt/1pt]
     Qwen3-VL-8B&\multirow{5}{*}{Qwen3-VL-8B}&34.7&24.4&64.3&39.8&34.5&44.5&31.4&43.7&39.7&-\\
     SpaceEra & 
& \cellcolor{lightgreen}41.1
& \cellcolor{lightgreen}27.9
& \cellcolor{lightgreen}69.3
& \cellcolor{lightgreen}46.3
& \cellcolor{lightgreen}35.0
& \cellcolor{lightgreen}47.0
& \cellcolor{lightgreen}37.0
& \cellcolor{lightgreen}50.2
& \cellcolor{lightgreen}44.2
& \cellcolor{lightgreen}\textbf{\color{darkgreen}{$\uparrow$ 4.5\%}} \\

     % \rowcolor{lightgreen}SpaceEra&&41.1&27.9&69.3&46.3&35.0&47.0&37.0&50.2&44.2&\textbf{{\color{darkgreen} $\uparrow$ 4.5\%}}\\
               +ScenePick&&41.9&29.8&71.1&46.9&35.7&49.4&38.5&50.7&45.6&\textbf{{\color{darkgreen} $\uparrow$ 1.4\%}}\\
     +SpaceAlign&&47.4&28.6&73.1&51.4&39.5&54.2&41.8&57.9&49.2&\textbf{{\color{darkgreen} $\uparrow$ 5.0\%}}\\
     SpaceEra++ & 
& \cellcolor{green!15!white}48.6
& \cellcolor{green!15!white}31.8
& \cellcolor{green!15!white}75.0
& \cellcolor{green!15!white}51.9
& \cellcolor{green!15!white}42.0
& \cellcolor{green!15!white}57.1
& \cellcolor{green!15!white}44.7
& \cellcolor{green!15!white}59.0
& \cellcolor{green!15!white}51.3
& \cellcolor{green!15!white}\textbf{\color{darkgreen}{$\uparrow$ 7.1\%}} \\

     % \rowcolor{green!15!white}SpaceEra++&&45.2&28.9&71.2&51.2&40.1&52.3&39.7&53.6&47.8&\textbf{{\color{darkgreen} $\uparrow$ 3.6\%}}\\

    \bottomrule
  \end{tabular}
    % \vspace{-2ex}
  \label{tab:vsi-bench}
\end{table*}

\textbf{Relative Relation Reward}. While the absolute coordinate reward $R_{\mathrm{abs}}$ evaluates the per-object localization accuracy, it is insensitive to the structural layout among objects. We therefore introduce a relative relation reward $R_{\mathrm{rel}}$ that measures the consistency between predicted and ground-truth pairwise spatial relations.

Let $\mathbf{p}_i=(x_{p,i},y_{p,i})$ and $\mathbf{g}_i=(x_{g,i},y_{g,i})$ denote the predicted and ground-truth coordinates of the $i$-th object, respectively. For any ordered pair $(i,j)$ with $i\neq j$, we define the relation vectors $\mathbf{r}^{(p)}_{ij}=\mathbf{p}_j-\mathbf{p}_i$ and $\mathbf{r}^{(g)}_{ij}=\mathbf{g}_j-\mathbf{g}_i$. To compare relations, we consider both directional consistency and distance-ratio consistency.

\paragraph{Directional consistency}
We quantify the alignment between the two relation vectors using cosine similarity:
\begin{equation}
\mathrm{cos\_sim}_{ij}
=\frac{\mathbf{r}^{(p)}_{ij}\!\cdot\!\mathbf{r}^{(g)}_{ij}}
{\|\mathbf{r}^{(p)}_{ij}\|_2\,\|\mathbf{r}^{(g)}_{ij}\|_2+\varepsilon},
\label{eq:cosine}
\end{equation}
where $\varepsilon>0$ avoids division by zero. The corresponding angular discrepancy is:
\begin{equation}
L_{\mathrm{ang},ij}=1-\mathrm{cos\_sim}_{ij}\in[0,2].
\label{eq:angle-loss}
\end{equation}

\paragraph{Distance-ratio consistency}
To preserve relative spacing, we penalize the normalized deviation between the predicted and ground-truth pairwise distances:
\begin{equation}
L_{\mathrm{dist},ij}
=\frac{\big|\,\|\mathbf{r}^{(p)}_{ij}\|_2-\|\mathbf{r}^{(g)}_{ij}\|_2\,\big|}
{\sqrt{2}\,M}.
\label{eq:dist-loss}
\end{equation}
% where $M$ is the grid size used for spatial normalization.

The pairwise relation discrepancy combines the two terms with a balance parameter $\alpha\in[0,1]$:
\begin{equation}
d_{ij}=\alpha\,L_{\mathrm{ang},ij}+(1-\alpha)\,L_{\mathrm{dist},ij}.
\label{eq:pairwise-discrepancy}
\end{equation}
We then aggregate over all ordered pairs to obtain the relative relation reward:
\begin{equation}
R_{\mathrm{rel}}
=1-\frac{1}{N(N-1)}\sum_{i\neq j} d_{ij}.
\label{eq:Rrel}
\end{equation}

\subsubsection{\textbf{Overall Reward}} Based on the above components, we present the final reward function:
\begin{equation}
R =
R_{\mathrm{format}} + R_{\mathrm{task}} + R_{\mathrm{abs}} + R_{\mathrm{rel}},
\end{equation}
where $\mathrm{task} \in \{\mathrm{mc}, \mathrm{num}\}$. The advantage $A_i$, measuring the relative quality of response $o_i$ within a group $[o_1, o_2, \cdots, o_G]$, is computed from the normalized rewards $A_i = \frac{R - \mathrm{mean}(\{R\})}{\mathrm{std}(\{R\})} $
To constrain the optimized policy $\pi_\theta$ from drifting too far from the reference model $\pi_{\mathrm{ref}}$, a KL regularization term $\mathcal{D}_{\mathrm{KL}}(\pi_\theta\|\pi_{\mathrm{ref}})$ is added. The final policy update follows the GRPO clipped surrogate objective:
\begin{align}
J(\theta)
&= \mathbb{E}_{q,\{o_i\}} \Bigg[
\frac{1}{G} \sum_{i=1}^{G}
\Big(
\min\!\Big(
\frac{\pi_\theta(o_i|q)}{\pi_{\mathrm{old}}(o_i|q)} A_i, \nonumber\\
&
\mathrm{clip}\!\Big(
\frac{\pi_\theta(o_i|q)}{\pi_{\mathrm{old}}(o_i|q)}, 1 - \epsilon, 1 + \epsilon
\Big) A_i
\Big)
\Big)
- \beta\, \mathcal{D}_{\mathrm{KL}}(\pi_\theta \| \pi_{\mathrm{ref}})
\Bigg],
\label{eq:grpo}
\end{align}
where $\beta$ controls the strength of KL regularization and $\epsilon$ determines the clipping range.
}

\section{SpatialMind Prompting Inference}
As shown in Fig.~\ref{fig:cot}, our SpatialMind prompting strategy consists of two main components: \textbf{1) Scene Decomposition}, where the 3D scene depicted in the video is transformed into multiple different representations; and \textbf{2) Question Decomposition}, in which the question is broken down into a sequence of fine-grained reasoning steps.

\subsection{Scene Decomposition}
The scene decomposition process includes three sequential steps: local modeling, coordinate mapping, and cognition generation.

\subsubsection{\textbf{Local Modeling}}
The first step processes scanning video frames to extract object instances and their relative spatial configurations within localized coordinate systems. To handle scene complexity and reduce the search space, we leverage GPT-4o to identify all objects mentioned across the questions associated with a given scene, using them as candidate targets. For each frame \(i\), we prompt VLMs to detect a subset of objects \( \{c_{ij}\} \) from the candidate targets and estimate their positions \( \mathbf{p}_{ij}^{\mathrm{local}} \in \mathbb{R}^3 \). These positions are defined relative to a randomly selected reference object (i.e., origin) within the same frame, forming a local 3D map $\mathcal{L}_i = \left\{ (c_{ij}, \mathbf{p}_{ij}^{\mathrm{local}}) \mid j = 1, \dots, n_i \right\}$,
% \begin{equation}
%     \mathcal{L}_i = \left\{ (c_{ij}, \mathbf{p}_{ij}^{\text{local}}) \mid j = 1, \dots, n_i \right\},
% \end{equation}
where $n_i$ denotes the number of objects in frame $i$. Because each video frame captures only a limited field of view, the same object may appear across multiple frames from different perspectives. Thus, this step focuses on accurate per-frame object detection and spatial localization, laying the foundation for subsequent alignment in a global coordinate system.

% This step captures the egocentric spatial layout of objects across frames, effectively segmenting and localizing instances under various viewpoints.
%Importantly, since each video frame reflects a narrow field of view, the same object may appear in multiple frames under different perspectives. Local modeling therefore emphasizes accurate per-frame instance detection and spatial anchoring, which later enables alignment in the global coordinate system.

\subsubsection{\textbf{Coordinate Mapping}}
To integrate spatial information across video frames, this step transforms all locally detected object positions into a unified global coordinate system. The global origin is defined by selecting the reference object in the first frame.
To estimate motion between frames, we prompt the VLM to infer the relative rotation and translation between adjacent frames. These relative transformations are accumulated sequentially to compute each frame’s transformation $\mathbf{T}_i$ with respect to the global coordinate system:
\begin{equation}
    \mathbf{T}_i = \prod_{k=1}^{i}
\begin{bmatrix}
\mathbf{R}_{k,k-1} & \mathbf{t}_{k,k-1} \\
\mathbf{0} & 1
\end{bmatrix},
\end{equation}
where $\mathbf{R}_{k,k-1}$ and $\mathbf{t}_{k,k-1}$ denote the relative rotation and translation from frame $k-1$ to frame $k$, respectively. This accumulated approach provides more stable and accurate alignment than directly estimating each frame’s absolute pose. Using these transformations, each object’s local coordinates are converted into global coordinates via homogeneous transformation: 
\begin{equation}
    \begin{bmatrix}
\mathbf{p}_{ij}^{\mathrm{global}} \\
1
\end{bmatrix}
=
\mathbf{T}_i \cdot
\begin{bmatrix}
\mathbf{p}_{ij}^{\mathrm{local}} \\
1
\end{bmatrix},
\end{equation}
where $\mathbf{p}_{ij}^{\mathrm{global}}$ denotes the global coordinates of the object $j$ in the frame $i$. This step ensures that all detected objects across frames are positioned consistently within the same 3D space. Since objects may appear in multiple frames under different perspectives, we merge duplicate detections based on spatial proximity and semantic consistency via prompting. The result is a global 3D map of the scene $\mathcal{G} = \left\{ (c_k, \mathbf{p}_k^{\mathrm{global}}) \right\}_{k=1}^{N}$,
% \begin{equation}
%     \mathcal{G} = \left\{ (c_k, \mathbf{p}_k^{\text{global}}) \right\}_{k=1}^{N},
% \end{equation}
where $N$ is the total number of all object instances in the entire scene. This map serves as a unified spatial abstraction that captures the overall layout from egocentric scanning videos.

\subsubsection{\textbf{Cognition Generation}}
Beyond constructing a 3D map, we explore two additional formats for representing scene structure: a 2D spatial grid and natural language descriptions. 
% With a complete global map, the system proceeds to generate a structured spatial representation for reasoning.
We define a regular 2D grid over the global scene, typically aligned with the \(XY\)-plane. Each grid cell corresponds to a fixed real-world area (e.g., 1 meter per cell, denoted by cell size \(s\) ). Each object \(c_k\) is mapped to a discrete grid location $ (i_k, j_k) = \left( \left\lfloor \frac{x_k}{s} \right\rfloor, \left\lfloor \frac{y_k}{s} \right\rfloor \right)$,
% \begin{equation}
%     (i_k, j_k) = \left( \left\lfloor \frac{x_k}{s} \right\rfloor, \left\lfloor \frac{y_k}{s} \right\rfloor \right),
% \end{equation}
where \( (x_k, y_k) \) are the horizontal components of the object’s global position \( \mathbf{p}_k^{\mathrm{global}} \). In parallel, we generate natural language descriptions of object locations relative to a designated reference point. Using prompting, the model produces statements such as \{``monitor'': ``locate 1 meter to the left of the reference point''\}. These descriptions serve as a human-interpretable form of spatial cognition, bridging visual perception and symbolic reasoning. 

% These spatially grounded outputs, combined with egocentric video, are finally passed to VLMs to support downstream tasks like spatial question answering. This cognitive generation step enables abstract, language-level reasoning grounded in concrete visual observations and structured spatial priors.

\subsection{Question Decomposition}
Different types of spatial questions require distinct reasoning strategies. To accommodate this diversity, we first categorize questions into several types (e.g., object size, relative distance, and relative direction). For each category, we design a dedicated reasoning procedure using GPT-4o, followed by human verification to ensure correctness and interpretability. For instance, consider a question from the ``relative distance'' category: \textit{Among the refrigerator, window, and microwave, which object is closest to the door?} The reasoning process for this type follows four structured steps: 1) Identify all mentioned objects, 2) Estimate the spatial coordinates of all relevant objects, 3) Compute the pairwise distances between the door and each candidate object,  and 4) Select the object with the minimum distance as the answer. During inference, the system correspondingly selects the appropriate reasoning procedure based on the identified question type.

To perform 3D spatial reasoning, we feed the VLMs with the input scanning video, one form of scene representation (e.g., 3D map, 2D grid, or textual position descriptions), the question, and the corresponding step-by-step reasoning plan. 
% \begin{table}[t]
% \centering
% \caption{Comparison of different benchmarks. Note that the statistics for the ScanQA and SQA3D datasets are reported on their respective validation sets.}
%   % \renewcommand{\arraystretch}{1.1} % 默认是1.0，1.5表示增加50%行高
% {
% \begin{tabular}{M{1.5cm}M{1.5cm}M{1.5cm}M{2cm}}
% \toprule
% \textbf{Dataset} & \textbf{\#Questions} & \textbf{\#Scenes} & \textbf{Source} \\
% \midrule
% VSI-Bench & $>$5,000 & 288 & ScanNet, ScanNet++, ARKitScenes \\
% \midrule
% OpenEQA  & 1,899 & $>$180 & ScanNet, HM3D \\
% \midrule
% ScanQA    & 9,353   & 71 & ScanNet \\
% \midrule
% SQA3D     & 3,261   & 65 & ScanNet \\
% \bottomrule
% \end{tabular}
% \label{benchmark_compare}
% }
% \end{table}

\section{Experiments}
{
\subsection{Experimental Settings}
\subsubsection{Benchmarks}
We conducted evaluations on four comprehensive benchmarks, including VSI-Bench~\cite{yang2024thinking}, OpenEQA~\cite{majumdar2024openeqa}, ScanQA~\cite{azuma2022scanqa}, and SQA3D~\cite{masqa3d}. They span a range of spatial tasks, such as object recognition, attribute estimation, spatial understanding, and functional reasoning. 
% A detailed comparison of different benchmarks is shown in Table~\ref{benchmark_compare}.

\subsubsection{Baselines}
As shown in Tables~\ref{tab:vsi-bench}, our evaluation covers a diverse set of VLM baselines spanning multiple architectures and scales. These include closed-source models such as Gemini-2.0 Flash\footnote{\url{https://deepmind.google/technologies/gemini/flash/}.}, Gemini-1.5 Pro\footnote{\url{https://deepmind.google/technologies/gemini/pro/}.}, and GPT-4o~\cite{hurst2024gpt}, as well as open-source models, encompassing both spatial-specific models (e.g., Struct2D~\cite{zhu2025struct2d}, Spatial-MLLM~\cite{wu2025spatial}, and SpaceVista~\cite{sun2025spacevista}) and general-purpose models (e.g., InternVL2~\cite{chen2024expanding} with 8B and 40B parameters, Qwen2.5-VL~\cite{bai2025qwen2} with 7B and 72B parameters, and Qwen3-VL~\cite{bai2025qwen3} with 8B parameters). 

\subsubsection{Configurations}
% 使用数据
Gemini models operate at a fixed sampling rate of 1 frame per second (FPS). InternVL processes 8 frames per video, whereas other models process 16 frames per video, all at a resolution of 224×224.
We set the global batch size to 16 and used a linear learning-rate schedule, with a peak value of $10^{-5}$.
For ScenePick sampling, we set $N_m=128$ and $N_s=16$. The hyperparameter $\lambda$ is 20.
In the SpaceAlign training, we performed 8 rollouts per question and set the default sampling temperature to 1. The parameter $M$ is set to 10 for the VSI-Bench dataset and to 8 for all other datasets. The coefficient $\varepsilon$, $\alpha$, and $\beta$ is set to 0.1, 0.6, and 0.04, respectively. The hyperparameter $\epsilon$ is fixed to 0.2.
All experiments are conducted on 8 NVIDIA H20 GPUs.
}

% \begin{table}[t]
%   \caption{The effect among different components on VSI-Bench.}
%   \centering
%   {
%   \begin{tabular}{ccccc}
%     \toprule
%     \textbf{SpatialMind} & \textbf{ScanForgeQA} & \textbf{ScenePick}&\textbf{SpaceAlign}&\textbf{Avg} \\
%     \midrule
%     &&&&37.2\\
%     \ding{51}&\ding{51}&&&43.9\\
%     \ding{51}&&\ding{51}&\\
%     \ding{51}&&&\ding{51}&\\
%     \ding{51}&\ding{51}&\ding{51}&\\
%     \ding{51}&&\ding{51}&\ding{51}&\\
%     \bottomrule
%   \end{tabular}}
%   \label{tab:each_other}
% \end{table}

\begin{table}[t]
\centering
\caption{Performance comparison on the EM-EQA subset of OpenEQA and the validation set of ScanQA and SQA3D. The backbone is Qwen2.5-VL-7B.}
\begin{tabular}{lccc}
\toprule
\multirow{2}{*}{\textbf{Method}} & \textbf{OpenEQA}& \textbf{ScanQA}& \textbf{SQA3D}\\

&Acc/Score & BLEU-1&EM-1\\
\midrule
Qwen2.5-VL-7B& 50.1/3.1&32.5&17.2\\
       SpaceEra&58.6/3.5&37.9&24.5\\

          +ScenePick&59.5/3.5&40.1&26.3\\
         +SpaceAlign&61.1/3.6&44.2&28.5\\ 
         SpaceEra++&63.6/3.6&46.5&30.5\\
% \cdashline{1-4}[1pt/1pt]
% Qwen3-VL-8B& \\
%      \rowcolor{lightgreen}Qwen3-VL-8B (v1)&\\

%           +ScenePick\\
%          +SpaceAlign\\ 
%          \rowcolor{green!15!white}Qwen3-VL-8B (v2)\\

\bottomrule
\end{tabular}
    \label{tab:openeqa}
\end{table}

% \begin{table}[t]
%     \centering
% \caption{Effects of different training strategies.}
% \begin{tabular}{lcc}
% \toprule
% \textbf{Method} & \makecell{\textbf{Room} \\ \textbf{Size}}&\textbf{Avg} \\
% \midrule
% Qwen2.5-VL-7B&38.9&37.2\\
% +SFT\\  
% +RL\\ 
% \bottomrule
% \end{tabular}
%     \label{tab:training}
% \end{table}

\subsection{Performace Comparison}
We investigated the following multiple key questions to assess our approach:

% We reported the results of our method and fine-tuning data across different model architectures, parameter scales, and evaluation datasets, as shown in Tables~\ref{tab:vsi-bench} and~\ref{tab:openeqa}. Detailed experimental analysis is presented below:
% \subsubsection{\textbf{Which scene representation format is most interpretable by traditional VLMs?}}
% Fig.~\ref{fig:format_compare}  presents a performance comparison across different representation formats: no additional spatial context (Base), inclusion of a 3D map (+Map), a 2D grid (+Grid), and object-centric textual descriptions (+Des). Across all models, a consistent trend emerges: the +Des variant outperforms others, followed by +Grid, while +Map yields the least improvement. These results suggest that current VLMs are more adept at interpreting one-dimensional textual descriptions than high-dimensional structured spatial formats. Consequently, we adopted the textual description in subsequent experiments as the default scene representation. 

\begin{table}[t]
    \centering
\caption{Effects of different frame sampling approaches and RL training strategies on VSI-Bench. The backbone is Qwen2.5-VL-7B.}
\begin{tabular}{clc}
\toprule
\textbf{Row}&\textbf{Method} &\textbf{Avg} \\
\midrule
1&SpaceEra&43.9\\
2&+ScenePick (w/o object semantics)&44.9\\  
3&+ScenePick (w/o spatial coverage)&44.3\\ 
4&+ScenePick&45.2\\
  
5&+SpaceAlign (w/o space imagination reward) &45.8\\  
6&+SpaceAlign (w/o absolute coordinate reward) &46.3\\ 
7&+SpaceAlign (w/o relative relation reward) &47.9\\ 
8&+SpaceAlign&48.7\\
9&SpaceEra++&50.3\\
\bottomrule
\end{tabular}
    \label{tab:ablation}
\end{table}

\begin{table}[t]
\centering
\caption{{Multi-seed robustness analysis of the added components on VSI-Bench and the backbone is Qwen3-VL-8B.}}
\label{tab:robustness}
\begin{tabular}{lcc}
\toprule
\textbf{Method} & \textbf{Avg} & $\Delta$ \\
\midrule
Base                  & 39.7                & - \\
SpaceEra              & 44.2                & $\uparrow$ 4.5\% \\
\midrule
+ScenePick            & 45.6                & \textbf{\color{darkgreen}{$\uparrow$ 1.4\%}} \\
\midrule
+SpaceAlign (Seed 42)  & 49.2                & \textbf{\color{darkgreen}{$\uparrow$ 5.0\%}} \\
+SpaceAlign (Seed 0)  & 48.9                & \textbf{\color{darkgreen}{$\uparrow$ 4.7\%}} \\
+SpaceAlign (Seed 3407)  & 49.1                & \textbf{\color{darkgreen}{$\uparrow$ 4.9\%}} \\
\textbf{+SpaceAlign (Mean $\pm$ Std)} & \textbf{49.1 $\pm$ 0.2} &-\\
\bottomrule
\end{tabular}
\end{table}

\begin{table}[t]
\centering
\caption{{Sensitivity analysis of GPT-4o-based decomposition in the SpatialMind prompting strategy on VSI-Bench.}}
\label{tab:gpt4o_replace}
\begin{tabular}{lccc}
\toprule
\textbf{Backbone} & \textbf{Recognizer} & \textbf{Avg} & $\Delta$ \\
\midrule
\multirow{4}{*}{Qwen2.5-VL-72B} 
& None        & 39.2 & - \\
& GPT-4o        & 44.0 & \textbf{\color{darkgreen}{$\uparrow$ 4.8\%}} \\
& Qwen3-VL-8B   & 42.3 & \textbf{\color{darkgreen}{$\uparrow$ 3.1\%}} \\
& Qwen3-VL-30B  & 43.5 & \textbf{\color{darkgreen}{$\uparrow$ 4.3\%}} \\
\bottomrule
\end{tabular}
\end{table}

{
\subsubsection{\textbf{How do ScenePick and SpaceAlign impact VLM performance?}}
As shown in Table~\ref{tab:vsi-bench}, we independently introduced the ScenePick frame sampling and the SpaceAlign training strategy across a range of VLMs. The results reveal three key findings:
a) Both ScenePick and SpaceAlign consistently improve visual-spatial understanding across models. This includes large-scale proprietary models such as Gemini-1.5 Pro and the open-source models enhanced in the conference version, demonstrating the effectiveness and generalizability of the extended methods. 
b) Finetuning the model with SpaceAlign strategy yields larger gains compared to ScenePick sampling. For instance, Qwen2.5-VL-7B gains 4.8\% from SpaceAlign fine-tuning, compared to only 1.3\% from ScenePick sampling.
c) Humans and VLMs exhibit complementary strengths. Human participants excel in qualitative tasks (e.g., achieving 100\% accuracy on the \textit{Appearance Order} task) but perform poorly on precise quantitative estimations (e.g., \textit{Object Size}). In contrast, VLMs show strong quantitative reasoning ability and, in some cases, even surpass human-level performance. This contrast underscores the potential of VLMs to complement human perception in spatial tasks.

\subsubsection{\textbf{Can combining two components yield further gains?}}
To assess whether ScenePick and SpaceAlign provide complementary benefits, we concatenated these two components. The results, reported in the row marked ``SpaceEra++'' in Table~\ref{tab:vsi-bench}, demonstrate consistent performance improvements across all evaluated models, confirming that the two approaches are complementary and can lead to further performance gains.
}

% \begin{figure}[t]
%   \centering
%     \includegraphics[width=0.7\linewidth]{img/mvbench.pdf}
%     \caption{Performance of Qwen2.5-VL-7B on MVBench and Video-MME.}
%     \label{fig:mvbench}
% \end{figure}

{
\subsubsection{\textbf{Does the improvement generalize to other spatial benchmarks?}}
To assess the generalizability of the extended methods, we conducted evaluations on multiple other benchmarks, including OpenEQA, ScanQA, and SQA3D.
As shown in Table~\ref{tab:openeqa}, both ScenePick sampling and SpaceAlign training lead to consistent performance gains across all benchmarks. These results further validate the standalone and combined advantages of our newly proposed method and confirm its applicability across diverse spatial tasks and datasets. 
%provides further evidence of the effectiveness of our prompting strategy and fine-tuning data, as reflected in consistently superior performance across a range of spatial reasoning benchmarks. These results substantiate the robustness and generalizability of our framework across diverse data sources and task categories.

% \subsubsection{\textbf{Does RL affect performance on other tasks?}}
% To investigate whether enhancing visual-spatial capabilities via fine-tuning adversely impacts a model's general performance,  we conducted evaluations on MVBench~\cite{li2024mvbench} and Video-MME~\cite{fu2024video}, two broad multi-task video benchmarks. As shown in Figure~\ref{fig:mvbench}, fine-tuning with ScanForgeQA slightly improves performance on MVBench but leads to a marginal drop on Video-MME. This difference likely stems from MVBench containing spatial reasoning tasks, while Video-MME focuses more on event comprehension. To mitigate this trade-off, we further experimented with mixed fine-tuning, combining a small proportion (5\% and 10\%) of traditional data from ShareGPT4Video~\cite{chen2024sharegpt4video} with ScanForgeQA. Results show that this strategy achieves improved performance, surpassing the original Qwen2.5-VL-7B baseline, suggesting that spatial fine-tuning can be harmonized with broader capabilities through data balancing. 
%The results are promising, with mixed fine-tuning achieving enhanced performance that surpasses the original Qwen2.5-VL-7B model.
}

\subsubsection{\textbf{How is the answer to the metric-based question determined?}}
Metric-scale estimation is supported by the joint effect of several factors: a) Semantic Priors. Semantic priors play an important role, since many indoor object categories, such as chairs, tables, doors, and beds, typically fall within relatively stable size ranges. b) Contextual Reference. Contextual references from surrounding objects and the overall scene layout provide useful anchors for relative scale estimation. c) Multi-view Calibration. Compared with a single image, a scanning video offers richer multi-view cues, including viewpoint changes, parallax, occlusion variation, and broader scene coverage, which help reduce scale ambiguity.
Accordingly, the reported performance on metric-based questions should be interpreted as the result of multi-view visual evidence combined with learned semantic priors, rather than pure geometry alone.

\subsubsection{\textbf{Is the gain from the newly added module robust?}}
To further assess the robustness of the performance gains introduced by the newly added component, we report repeated-run results for the +SpaceAlign variant using three different random seeds in Table~\ref{tab:robustness}. We choose this setting because ScenePick is a deterministic frame-selection strategy, while the principal source of stochasticity in our framework lies in the training process of SpaceAlign. As such, repeated runs of +SpaceAlign provide a direct evaluation of the stability of the main trainable component. The results show only marginal variation across different runs, and the observed gains remain consistent, confirming the robustness of the improvements brought by SpaceAlign.

% \begin{figure}[h]
% \centering
% \begin{minipage}[t]{0.38\textwidth}
%   \centering
%   \includegraphics[width=\textwidth]{img/mvbench.pdf} % 替换为你的图像路径
%   \caption{The results on MVBench.}
%   \label{fig:example}
% \end{minipage}
% \hfill
% \begin{minipage}[t]{0.48\textwidth}
%   \centering
%   \captionof{table}{Ablation study on Qwen2.5-VL-7B with different frames and resolution.}
%   \begin{tabular}{ccc}
%     \toprule
%     \textbf{Frame} & \textbf{Resolution} & \textbf{Avg} \\
%     \midrule
%     \multirow{3}{*}{8} & 128 \\
%      & 256 \\
%      & 512 \\
%      \midrule
%     16 & 512 &33.1\\
%     \midrule
% \rowcolor{gray!20}    32 & 512 &37.2\\
%     \bottomrule
%   \end{tabular}
%   \label{tab:example}
% \end{minipage}
% \end{figure}

% \begin{figure}[t]
%     \centering
%     \includegraphics[width=0.7\linewidth]{img/gain.pdf}
%     \caption{Parameter analysis under different frame counts and resolutions.}
%     \label{fig:ablation2}
% \end{figure}

\begin{figure}[t]
    \centering
    %
    % ---------- 子图 (a) ----------
    \begin{minipage}{0.49\linewidth}
        \centering
        \includegraphics[width=\linewidth]{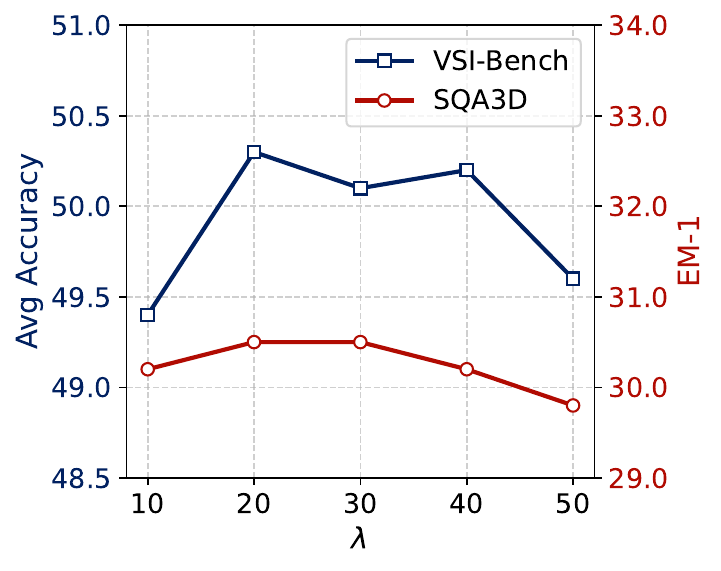}
        % \vspace{1pt}
        {\footnotesize (a) Parameter $\lambda$} % 手写子图标签，不会触发编号
    \end{minipage}
    \hfill
    %
    % ---------- 子图 (b) ----------
    \begin{minipage}{0.49\linewidth}
        \centering
        \includegraphics[width=\linewidth]{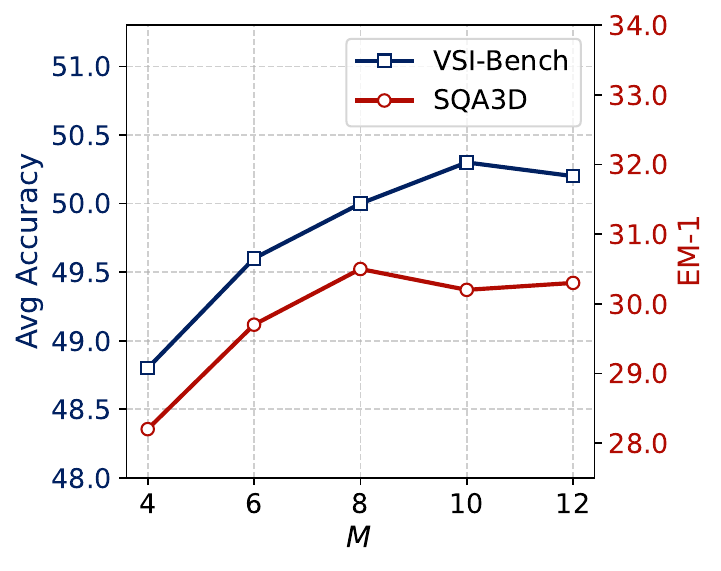}
        % \vspace{1pt}
        {\footnotesize (b) Parameter $M$}
    \end{minipage}
    \caption{Parameter analysis under different hyperparameters $\lambda$ and $M$.}
    \label{fig:ablation1}
\end{figure}

{
\subsection{Ablation Study}
In this section, we explored the impact of various design choices on the performance of SpaceEra++, using Qwen2.5-VL-7B as the backbone model, as shown in Table~\ref{tab:ablation}. 

\subsubsection{\textbf{On ScenePick sampling}}
To isolate the contributions of each component in the ScenePick frame sampling strategy, we evaluated three variants: 
\begin{itemize}
    \item (\textit{Row 2}): Ignoring object semantics and considering only scene-space coverage during selection, i.e., treating all 3D points as equally important.
    \item (\textit{Row 3}): Ignoring scene-space coverage and considering only the target objects, i.e., retaining only the 3D points belonging to the objects of interest.
    \item (\textit{Row 4}): Considering both factors simultaneously.
\end{itemize}

From the results, we observed that considering only a single factor is insufficient, leading to lower performance than jointly accounting for both. In addition, relying solely on spatial coverage outperforms using object semantics alone, likely because spatial coverage provides a more comprehensive representation of the scene’s geometry, whereas frames selected purely based on object semantics tend to be more concentrated.

\subsubsection{\textbf{On SpaceAlign training}}
To investigate the sources of effectiveness of our proposed SpaceAlign training strategy, we designed the following four variants:
\begin{itemize}
    \item (\textit{Row 5}): Removing the space imagination rewards (including $R_{\mathrm{abs}}$ and $R_{\mathrm{rel}}$) and optimizing the model using only rewards $R_{\mathrm{format}}$ and $R_{\mathrm{task}}$.
    \item (\textit{Row 6}): Removing the absolute coordinate reward $R_{\mathrm{abs}}$ and optimizing the model using rewards $R_{\mathrm{format}}$, $R_{\mathrm{task}}$, and $R_{\mathrm{rel}}$.
    \item (\textit{Row 7}): Removing the relative relation reward $R_{\mathrm{rel}}$ and optimizing the model using other three rewards.
    \item (\textit{Row 8}): Optimizing the model using all rewards, including $R_{\mathrm{format}}$, $R_{\mathrm{task}}$, $R_{\mathrm{abs}}$, and $R_{\mathrm{rel}}$.
\end{itemize}

From the results, we observed that removing all space imagination rewards leads to the largest performance drop, indicating that geometric guidance is essential for stable alignment. When the absolute coordinate reward $R_{\mathrm{abs}}$ is excluded, performance improves compared with \textit{Row 5} but still remains noticeably lower than the full model, suggesting that absolute spatial cues play an important role in grounding the model to global scene structure. Conversely, removing the relative relation reward results in a milder degradation, showing that relative positional reasoning provides complementary benefits. Finally, the full model using all reward components achieves the best performance, demonstrating that absolute and relative spatial signals jointly enhance the VLM's capability to align and interpret spatial information.

\begin{figure}[t]
    \centering
    \includegraphics[width=\linewidth]{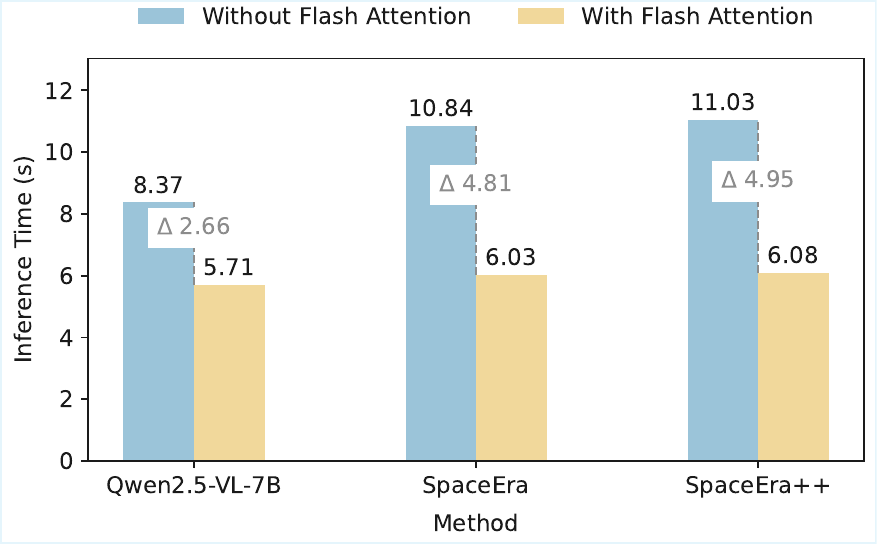}
    \caption{Inference time comparison with and without Flash Attention.}
    \label{fig:flash-attn-time}
\end{figure}

% \begin{table}[t]
%   \caption{Inference time comparison with and without Flash Attention.}
%   \centering
%   {
%   \begin{tabular}{lcc}
%     \toprule
%     \textbf{Method} & \textbf{Time} & \textbf{Time (with Flash Attention)} \\
%     \midrule
%     Qwen2.5-VL-7B & 8.37s & 5.71s \\
%     SpaceEra & 10.84s & 6.03s \\
%     SpaceEra++ & 11.03s & 6.08s\\
%     \bottomrule
%   \end{tabular}}
%   \label{tab:flash-attn-time}
% \end{table}

\begin{figure}[t]
    \centering
    \includegraphics[width=\linewidth]{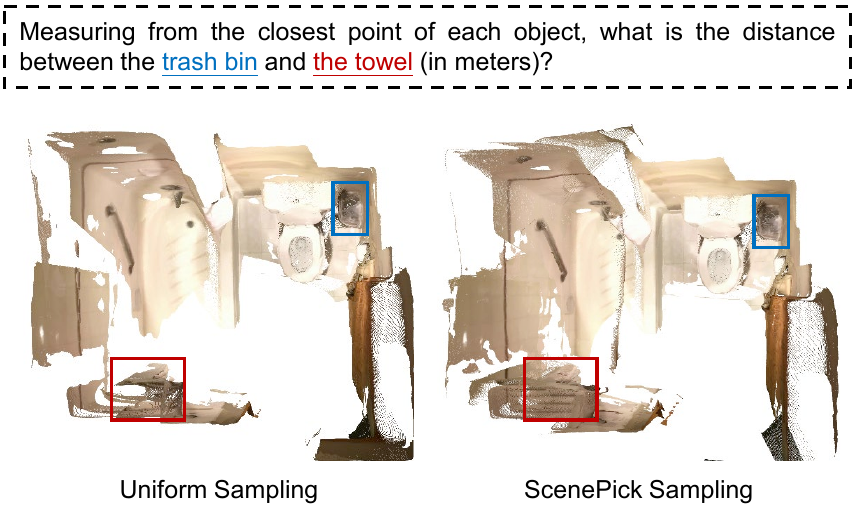}
    \caption{Comparison of scene using different frame sampling methods.}
    \label{fig:sampling_vis}
\end{figure}

\begin{figure}[t]
    \centering
    \includegraphics[width=\linewidth]{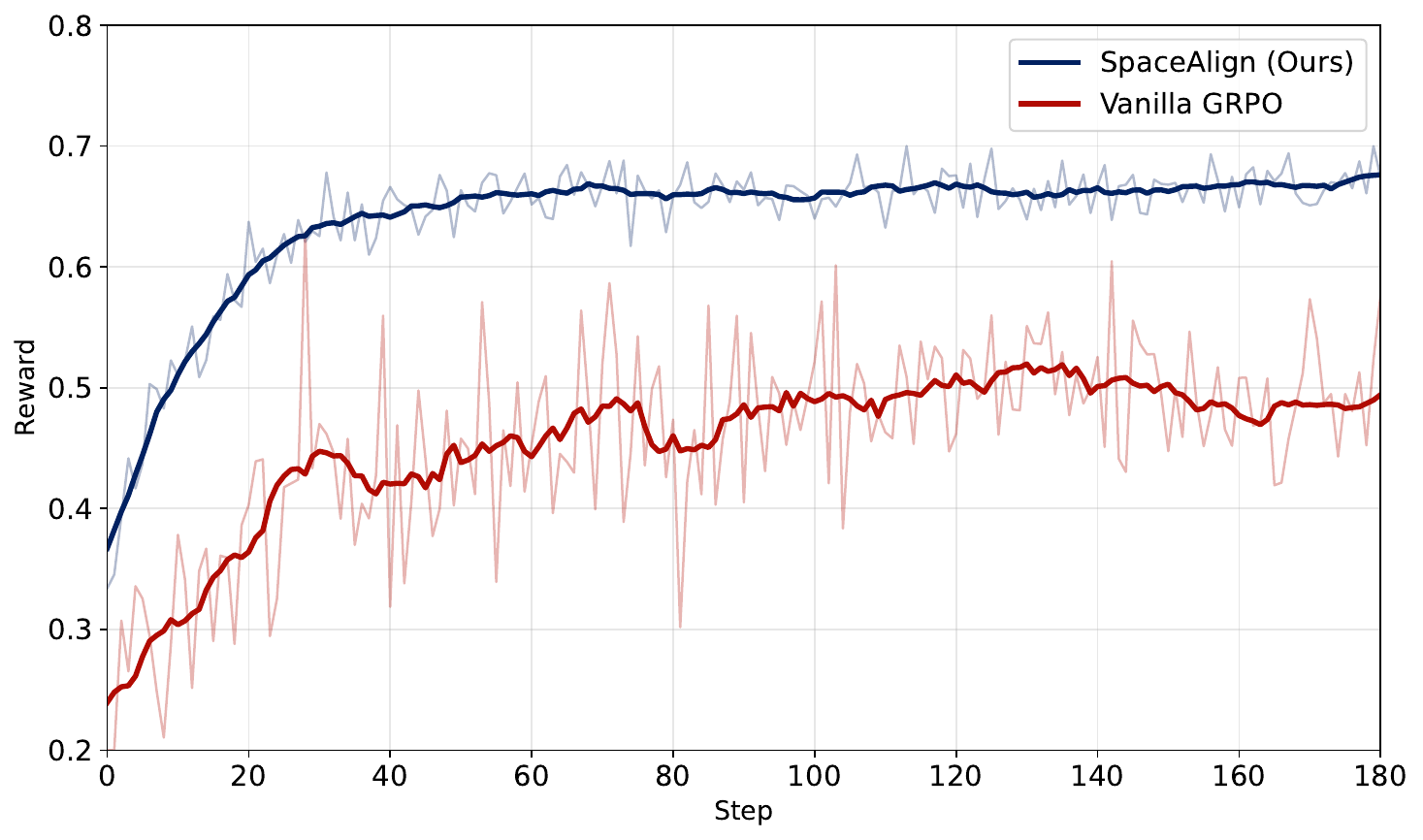}
    \caption{Reward curves along training steps under different RL strategies.}
    \label{fig:curve}
\end{figure}

\subsection{Parameter Analysis}
To evaluate the robustness of our approach, we analyzed its performance sensitivity to the voxel parameter $\lambda$ and grid size $M$, as illustrated in Fig.~\ref{fig:ablation1}. 

From the presented results, we derived the following findings:
\begin{itemize}
    \item As $\lambda$ increases, the voxel size $\Delta$ decreases, leading to a finer spatial discretization. This allows the coverage metric to capture more detailed variations in scene structure, enabling the selection of frames that provide richer and more diverse scene coverage. However, when $\lambda$ becomes excessively large, the voxels become overly fine, causing many voxels to contain only a few points or even noise. This may make the coverage estimation overly sensitive and potentially reduce its robustness.
    \item As $M$ increases, the spatial imagination grid becomes finer, enabling more detailed object distributions. However, the larger value reduces relative spatial errors, yielding higher and less discriminative rewards ($M$ appears in the denominator in Eqns. (13) and (17)). Although a finer grid offers richer spatial supervision, an excessively large $M$ can diminish reward sensitivity and introduce sparsity, potentially affecting training stability and final performance.
\end{itemize}
% 3) Our method consistently outperforms the baseline across all settings, with performance further improving as the number of frames and resolution increase. This indicates that our approach remains stable and effective under varying visual input conditions.
}

\begin{figure*}
  \centering
  \includegraphics[width=\linewidth]{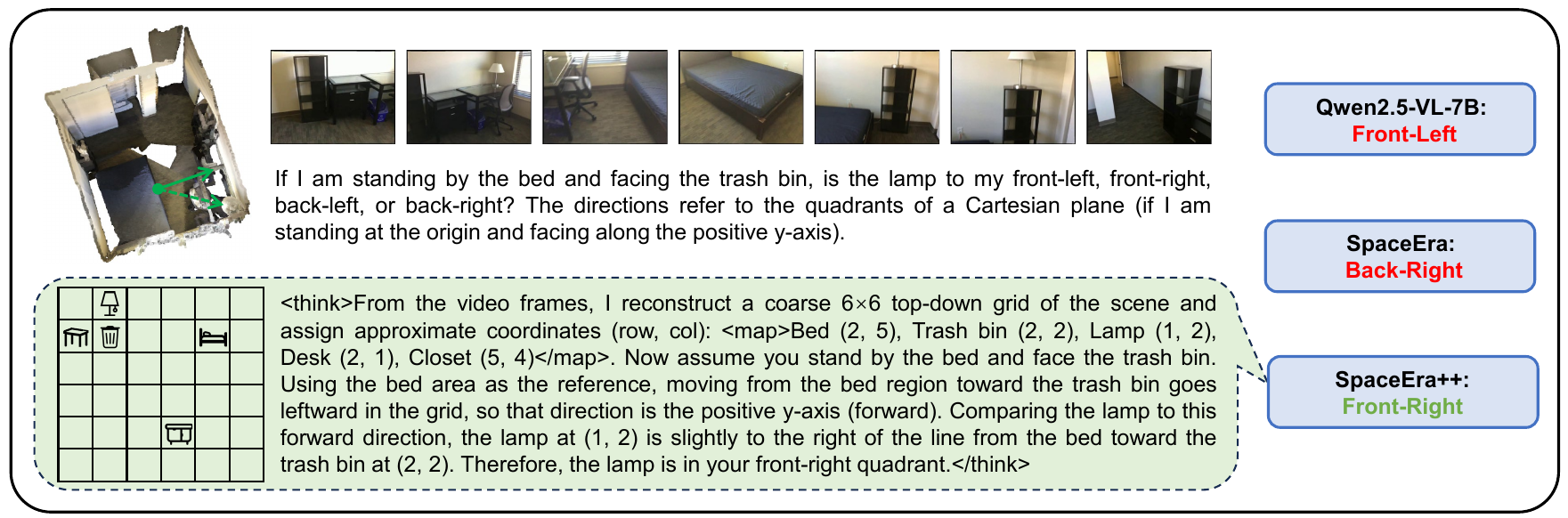}
  \caption{An example illustrating the thinking process of the SpaceAlign strategy, comparing predictions from Qwen2.5-VL-7B, the conference version SpaceEra, and the current extended version SpaceEra++. A 10×10 grid is used in practice, while a 6×6 grid is shown here solely for clearer visualization.}
  %\caption{Two cases from VSI-Bench.}
  \label{fig:case}
  % \vspace{-2ex}
\end{figure*}

% \begin{figure}[t]
%     \centering
%     %
%     % ---------- 子图 (a) ----------
%     \begin{minipage}{0.48\linewidth}
%         \centering
%         \includegraphics[width=\linewidth]{img/RL.pdf}
%         \vspace{2pt}
%         {\footnotesize (a) SFT} % 手写子图标签，不会触发编号
%     \end{minipage}
%     \hfill
%     %
%     % ---------- 子图 (b) ----------
%     \begin{minipage}{0.48\linewidth}
%         \centering
%         \includegraphics[width=\linewidth]{img/RL.pdf}
%         \vspace{2pt}
%         {\footnotesize (b) RLVR}
%     \end{minipage}
%     %
%     \caption{Curve visualization for different training strategies.}
%     \label{fig:curve}
% \end{figure}

\subsection{Sensitivity Analysis}
To evaluate the sensitivity of SpatialMind prompting strategy to the choice of GPT-4o, we replaced GPT-4o in both the scene decomposition and question decomposition stages with two open-source alternatives, namely Qwen3-VL-8B\footnote{\url{https://huggingface.co/Qwen/Qwen3-VL-8B-Instruct}.}
and Qwen3-VL-30B\footnote{\url{https://huggingface.co/Qwen/Qwen3-VL-30B-A3B-Instruct}.}, while keeping all other components unchanged. The corresponding results are reported in Table~\ref{tab:gpt4o_replace}.

The results show that the overall framework remains effective under both replacements, which verifies that the proposed method is not overly sensitive to the choice of GPT-4o and can be reproduced with fully open alternatives. The performance differences across substitutes are consistent with the capability gap of the underlying parser, but they do not affect the main conclusions of the paper.

{
\subsection{Inference Efficiency}
To evaluate the impact of SpaceEra++ on inference efficiency, we conducted experiments using the Qwen2.5-VL-7B as the backbone on an A100 GPU. Inference is performed on 100 samples, and we reported the average inference time per sample. The results are presented in Fig.~\ref{fig:flash-attn-time}.

As shown in the table, the extended methods do not introduce any noticeable increase in inference latency compared with the conference version model, referred to as SpaceEra, while delivering substantial performance improvements. In addition, applying inference acceleration techniques for VLMs, such as Flash Attention, can further reduce inference time and improve overall efficiency.
}

\subsection{Visualization}
\subsubsection{Sampling Rending}
To intuitively illustrate the differences in scene representation between uniform sampling and the proposed ScenePick sampling strategy, we rendered 16 video frames selected by each method using VGGT model, as shown in Fig.~\ref{fig:sampling_vis}. It can be observed that the scenes rendered from ScenePick sampling are more complete and exhibit richer structural details. Moreover, ScenePick places greater emphasis on key objects relevant to the query, such as the ``trash bin'' and the ``towel'', which are depicted with finer granularity, as highlighted by the solid bounding boxes. These results demonstrate that the frames selected by ScenePick provide higher-quality spatial structure, thereby enhancing the visual-spatial understanding capabilities of VLMs.

\subsubsection{Training Curves}
As shown in Fig.~\ref{fig:curve}, we presented the reward curves of our method under different training strategies (e.g., SpaceAlign and GRPO). Compared with Vanilla GRPO, SpaceAlign converges faster, achieves higher rewards, and shows smaller fluctuations, indicating more stable training and validating the effectiveness of the proposed spatial imagination mechanism.

\subsection{Qualitative Analysis}
Fig.~\ref{fig:case} illustrates a spatial reasoning case  on an indoor scene reconstructed from a scan, where the task is to determine the egocentric position of a lamp when standing by the bed and facing the trash bin. It compares different versions of Qwen2.5-VL-7B, showing that the conference version SpaceEra predicts an incorrect spatial relation, while the current journal version SpaceEra++ correctly identifies the lamp as being in the front-right direction. This improvement highlights that the journal version benefits from incorporating spatial imagination rewards during reasoning, which encourage the model to mentally construct a top-down spatial layout, perform viewpoint transformation, and reason over relative positions more systematically rather than relying on local visual cues.

\section{Conclusion}
In this work, we analyze the limitations of our NeurIPS 2025 Spotlight paper and propose novel solutions that extend it into a more comprehensive framework for visual-spatial understanding, termed SpaceEra++. Specifically, to address insufficient scene input, we introduce ScenePick, a frame sampling strategy that jointly considers scene coverage and object semantics to obtain a more complete scene representation without increasing the number of input frames. To further overcome limitations in spatial optimization, we design SpaceAlign, a spatially constrained GRPO algorithm that integrates both absolute and relative spatial cues to achieve more accurate spatial reasoning. Furthermore, we conduct extensive experiments across multiple benchmarks, demonstrating that these extensions consistently improve upon the original schema and thoroughly validate the effectiveness and generalization capability of the proposed framework.

{In the future, we will focus on more comprehensive evaluations in diverse real-world environments, including large-scale multi-room and outdoor scenarios, to further assess and improve the generalization of visual-spatial reasoning.}

\section*{Acknowledgments}
This work is supported by Shenzhen Science and Technology Program, No.:KQTD20240729102207002; Guangdong S\&T Program, No.:2025B0101130003; the National Natural Science Foundation of China, No.:62236003, No.:62376140, No.:62476071, No.:625B2065, No.:U24A20328, and No.:U23A20315;  Guangdong Basic and Applied Basic Research Foundation, No.:2025A1515011732; Beijing Natural Science Foundation, No.:4262074 and No.:L254018; the Science and Technology Innovation Program for Distinguished Young Scholars of Shandong Province Higher Education Institutions, No.:2023KJ128; the Special Fund for Taishan Scholar Project of Shandong Province; the Major Key Project of Pengcheng Laboratory, No.:PCL2025A14-4; the Fundamental Research Funds for the Central Universities, No.:HIT.DZJJ.2025048.

% {\appendix[Proof of the Zonklar Equations]
% Use $\backslash${\tt{appendix}} if you have a single appendix:
% Do not use $\backslash${\tt{section}} anymore after $\backslash${\tt{appendix}}, only $\backslash${\tt{section*}}.
% If you have multiple appendixes use $\backslash${\tt{appendices}} then use $\backslash${\tt{section}} to start each appendix.
% You must declare a $\backslash${\tt{section}} before using any $\backslash${\tt{subsection}} or using $\backslash${\tt{label}} ($\backslash${\tt{appendices}} by itself
%  starts a section numbered zero.)}

% %{\appendices
% %\section*{Proof of the First Zonklar Equation}
% %Appendix one text goes here.
% % You can choose not to have a title for an appendix if you want by leaving the argument blank
% %\section*{Proof of the Second Zonklar Equation}
% %Appendix two text goes here.}

% \section{References Section}
% You can use a bibliography generated by BibTeX as a .bbl file.
%  BibTeX documentation can be easily obtained at:
%  http://mirror.ctan.org/biblio/bibtex/contrib/doc/
%  The IEEEtran BibTeX style support page is:
%  http://www.michaelshell.org/tex/ieeetran/bibtex/
 
%  % argument is your BibTeX string definitions and bibliography database(s)
% %\bibliography{IEEEabrv,../bib/paper}
% You can manually copy in the resultant .bbl file and set second argument of $\backslash${\tt{begin}} to the number of references
%  (used to reserve space for the reference number labels box).

\bibliographystyle{IEEEtran}
\bibliography{reference.bib}

% \newpage
\begin{IEEEbiography}[{\includegraphics[width=1in,height=1.25in,clip,keepaspectratio]{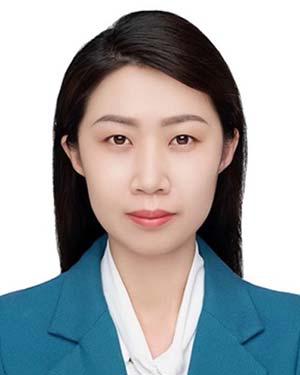}}]{Weili Guan}
received the master's degree from National University of Singapore, and the Ph.D. degree from Monash University. She has about 6 years of working experience at the enterprise. She is currently a professor at the School of Information Science and Technology, Harbin Institute of Technology (Shenzhen), China. Her research interests are multimedia computing and information retrieval. She has published more than 60 papers at the first-tier conferences and journals, like ACM MM, SIGIR, IEEE TPAMI and IEEE TIP.
\end{IEEEbiography}
\vspace{-2ex}
\begin{IEEEbiography}[{\includegraphics[width=1in,height=1.25in,clip,keepaspectratio]{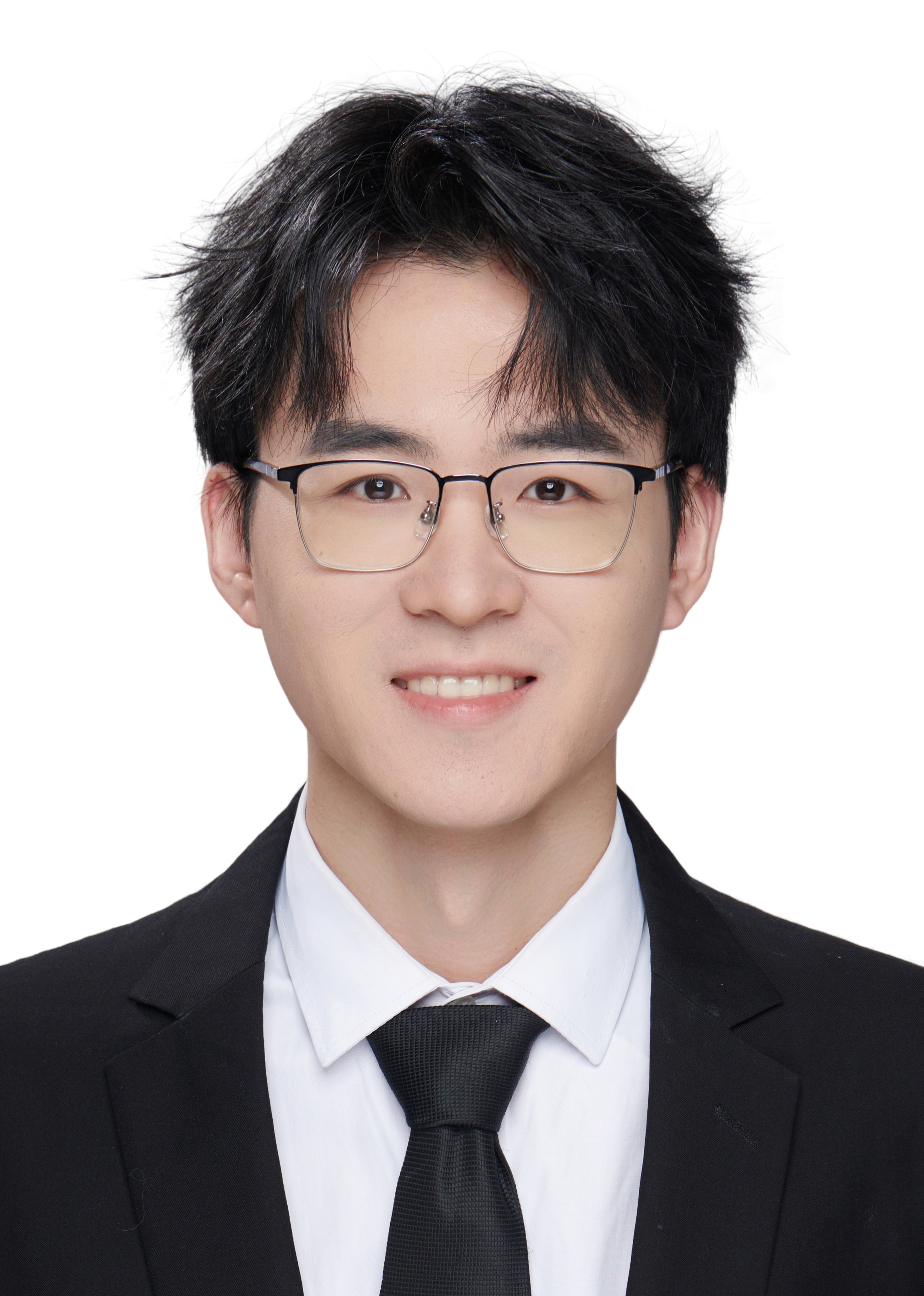}}]{Haoyu Zhang}
received the M.S. degree from Shandong University, in 2023. He is currently working toward the doctor's degree with the School of Computer Science and Technology, Harbin Institute of Technology (Shenzhen). His research interests include egocentric vision and spatial understanding. His work has been published in several top-tier conferences and journals, including IEEE TPAMI, CVPR, NeurIPS, ICML, AAAI, and ACM MM. He has served as a Reviewer for various conferences and journals, such as CVPR, ICCV, NeurIPS, ICML, ICLR, IEEE TPAMI, IEEE TKDE, and IEEE TMM.
\end{IEEEbiography}
\vspace{-2ex}
\begin{IEEEbiography}[{\includegraphics[width=1in,height=1.25in,clip,keepaspectratio]{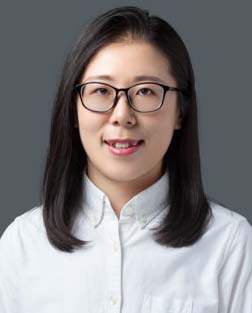}}]{Meng Liu}
received the M.S. degree in computational mathematics from the Dalian University of Technology, China, in 2016. She is currently a Professor with the School of Computer Science and Technology, Shandong Jianzhu University. Her research interests include multimedia computing and information retrieval. Various parts of her work have been published in top forums and journals, such as SIGIR, MM, and IEEE Transactions on Image Processing. She has served as a Reviewer and a Subreviewer for various conferences and journals, such as MMM 2018, ACM MM 2018/2019, JVCI, and INS.
\end{IEEEbiography}
\vspace{-2ex}
\begin{IEEEbiography}
[{\includegraphics[width=1in,height=1.25in,clip,keepaspectratio]{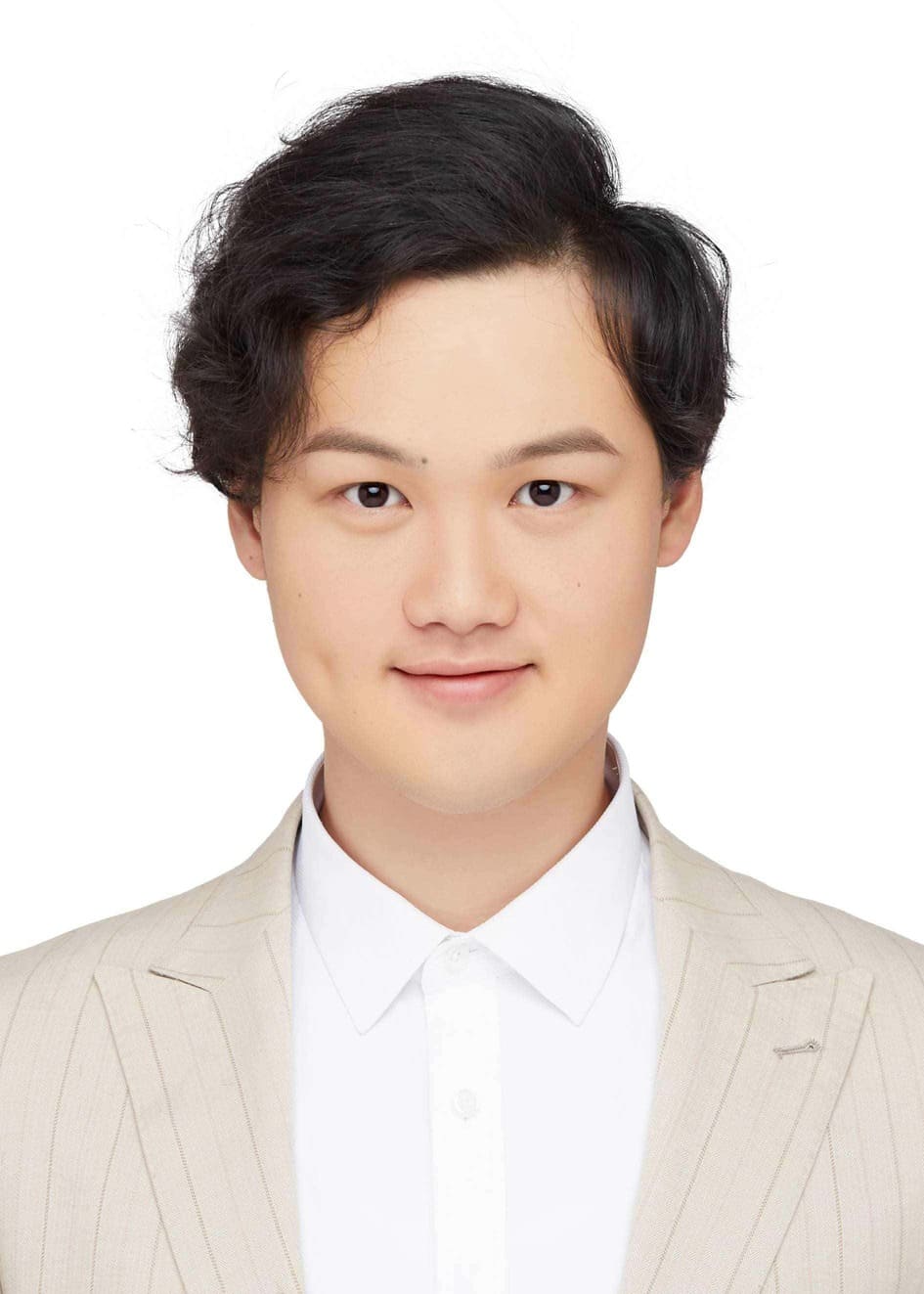}}]{Qianlong Xiang}
received the B.E. degree from the School of Computer Science and Techonology, Harbin Institute of Techonology, Weihai, China, in 2022. He is currently pursuing the Ph.D. degree at the School of Information Science and Techonology, Harbin Institute of Techonology, Shenzhen, China. His research has been published in top-tier conferences including CVPR. He has served as a reviewer for various conferences and journals, such as IEEE TPAMI, ACM MM and IEEE TCSVT. His main research interests include model compression and generative AI.
\end{IEEEbiography}
\vspace{-2ex}
\begin{IEEEbiography}[{\includegraphics[width=1in,height=1.25in,clip,keepaspectratio]{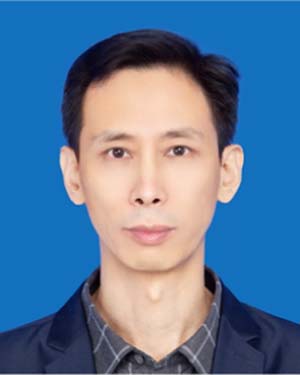}}]{Yaowei Wang}
received the PhD degree in computer science from Graduate University, Chinese Academy of Sciences, in 2005. He is currently an associate researcher with Pengcheng Laboratory, Shenzhen, China. He was an assistant professor with the School of Information and Electronics, Beijing Institute of Technology, and also was a guest assistant professor with the National Engineering Laboratory for Video Technology, Peking University, China. He has been the author or co-author of more than 50 refereed journals and conference papers. His research interests include machine learning and multimedia content analysis and understanding.
\end{IEEEbiography}
\vspace{-2ex}
\begin{IEEEbiography}[{\includegraphics[width=1in,height=1.25in,clip,keepaspectratio]{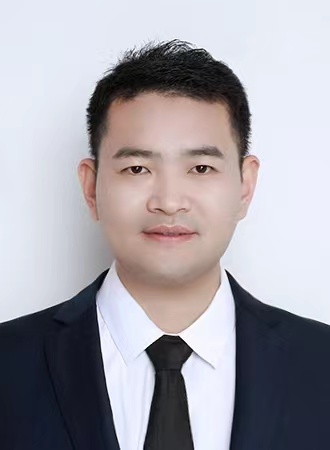}}]{Liqiang Nie} 
is currently the dean with the Department of Computer Science and Technology, Harbin Institute of Technology (Shenzhen). He received his B.Eng. and Ph.D. degree from Xi’an Jiaotong University and National University of Singapore (NUS), respectively. After PhD, Dr. Nie continued his research in NUS as a research fellow for three years. His research interests lie primarily in multimedia computing and information retrieval. Dr. Nie has co-/authored more than 100 papers and 4 books, received more than 34,000 Google Scholar citations. He is an AE of IEEE TKDE, IEEE TMM, IEEE TCSVT, ACM ToMM, and Information Science. Meanwhile, he is the regular area chair of ACM MM, NeurIPS, IJCAI and AAAI. He is a member of ICME steering committee. He has received many awards, like ACM MM and SIGIR best paper honorable mention in 2019, SIGMM rising star in 2020, TR35 China 2020, DAMO Academy Young Fellow in 2020, SIGIR best student paper in 2021, and ACM MM best paper in 2022.
\end{IEEEbiography}

\vfill

\end{document}